%% file: main.tex
\documentclass[lettersize,journal]{IEEEtran}
\usepackage{amsmath,amsfonts}
\usepackage{algorithmic}
\usepackage{algorithm}
\usepackage{colortbl}
\usepackage{array}
\usepackage[caption=false,font=normalsize,labelfont=sf,textfont=sf]{subfig}

\usepackage{textcomp}
\usepackage{stfloats}
\usepackage{url}
\usepackage{verbatim}

\usepackage{graphicx}
\usepackage{cite}
\usepackage{amssymb}
\usepackage{amsmath,mathrsfs,mathtools}
\usepackage[colorinlistoftodos]{todonotes}
\usepackage{xcolor}
\usepackage{comment}
\usepackage{tabularx,multirow}
\usepackage{hyperref}
\usepackage{xspace} 
\usepackage{subcaption} 
\usepackage{graphicx}
\usepackage{caption}
\usepackage{subcaption}

\usepackage[framemethod=TikZ]{mdframed}
\usepackage{adjustbox}

\renewcommand{\vec}[1]{\boldsymbol{\mathrm{#1}}}
\newcommand{\mtx}[1]{\boldsymbol{\mathrm{#1}}}

\newcommand{\mydef}{\ensuremath{\triangleq}}

\newcommand\E{\mathbb{E}}

\renewcommand{\vec}{\boldsymbol} 

\hypersetup{
   unicode=false,          
   pdftoolbar=true,        
   pdfmenubar=true,        
   pdffitwindow=false,     
   pdfstartview={FitH},    
   pdftitle={My title},    
   pdfauthor={Author},     
   pdfsubject={Subject},   
   pdfcreator={Creator},   
   pdfproducer={Producer}, 
   pdfkeywords={keyword1} {key2} {key3}, 
   pdfnewwindow=true,      
   colorlinks=false,       
   linkcolor=false,          
   citecolor=blue,        
   filecolor=magenta,      
   urlcolor=blue           
}

\mdfdefinestyle{MyFrame}{
    linecolor=black,
    outerlinewidth=.1pt,
    roundcorner=1pt,
    innertopmargin=\baselineskip,
    innerbottommargin=\baselineskip,
    leftmargin=1pt, 
    rightmargin=1pt, 
    backgroundcolor=gray!10!white
}

\begin{document}

\title{\textcolor{black}{Shortcut Learning in Binary Classifier Black Boxes: Applications to Voice Anti-Spoofing and Biometrics}}

\author{Md~Sahidullah$^{\dag}$,
        Hye-jin Shim$^{\dag}$,
        Rosa Gonzalez Hautam\"aki$^{\dag}$,
        and~Tomi~H.~Kinnunen$^{\dag}$
        \thanks{$^{\dag}$ Equal contribution by all authors.}
        \thanks{Corresponding author: Md Sahidullah.}
        \thanks{Md Sahidullah is with TCG CREST, Kolkata (e-mail: md.sahidullah@tcgcrest.org).}

\thanks{Hye-jin Shim is with Carnegie Mellon University, Pittsburgh, USA (e-mail: shimhz6.6@gmail.com).}

\thanks{Tomi Kinnunen is with School of Computing, University of Eastern Finland (UEF), FI-80101, Joensuu, Finland (e-mail: tomi.kinnunen@uef.fi).}

\thanks{Rosa Gonzalez Hautam\"aki is with University of Oulu, FI-90014, Oulu, Finland (e-mail: rosa.gonzalezhautamaki@oulu.fi), and University of Eastern Finland (UEF), FI-80101, Joensuu, Finland (e-mail: rgonza@uef.fi).}

}

\markboth{Submitted to IEEE Journal of Selected Topics in Signal Processing}%
{Shell \MakeLowercase{\textit{et al.}}: A Sample Article Using IEEEtran.cls for IEEE Journals}


\maketitle

\begin{abstract}
The widespread adoption of deep-learning models in data-driven applications has drawn attention to the potential risks associated with biased datasets and models. Neglected or hidden biases within datasets and models can lead to unexpected results. This study addresses the challenges of dataset bias and explores ``shortcut learning'' or ``Clever Hans effect'' in binary classifiers. We propose a novel framework for analyzing the black-box classifiers and for examining the impact of both training and test data on classifier scores. Our framework incorporates intervention and observational perspectives, employing a linear mixed-effects model for post-hoc analysis. By evaluating classifier performance beyond error rates, we aim to provide insights into biased datasets and offer a comprehensive understanding of their influence on classifier behavior. The effectiveness of our approach is demonstrated through experiments on audio anti-spoofing and speaker verification tasks using both statistical models and deep neural networks. The insights gained from this study have broader implications for tackling biases in other domains and advancing the field of explainable artificial intelligence.
\end{abstract}


\input{Section_Ia}

\input{Section_Ib}
\input{SECTION_III_Overview_of_Proposed_Framework}
\input{SECTION_METHODLOGY}
\input{SECTION_VI_Application_to_Antispoofing}

\input{SECTION_VII_Application_to_ASV}

\input{SECTION_DISCUSSION}

\input{SECTION_IX_Horse_Arithmetics_Broad_Outlook}
\input{SECTION_X_Conclusions}

\input{APPENDIX_Nuisance_feature_and_EER}

\section*{Acknowledgements}
This work was partially supported by Academy of Finland (Decision No. 349605, project ''SPEECHFAKES"). The authors thank the anonymous reviewers and the Associate Editor for their helpful comments and suggestions.


\bibliographystyle{IEEEtran}
\bibliography{mybib}

\end{document}

%% file: Section_Ia.tex
\section{Introduction}
\IEEEPARstart{T}{he} increasing adoption of deep-learning models in data-intensive applications~\cite{zue1990speech, doddington2000nist, nagrani2020voxceleb, wang2020asvspoof} has brought attention to risks associated with \emph{biased} datasets and models~\cite{torralba2011unbiased,liu2024decade}. Such biases, originating from flawed data or models, pose challenges for \textbf{reliable performance evaluation} and can lead to unexpected model behavior or undesirable system outcomes in real-world applications. Identifying these biases is essential, as hidden or overlooked biases can create significant challenges in machine learning systems. This issue, where models rely on unintended or spurious cues rather than meaningful patterns to solve problems, is known as \emph{shortcut learning}~\cite{geirhos2020shortcut}.

To deeply understand models and estimate inherent data bias, research on shortcut learning has recently gained attention~\cite{torralba2011unbiased, stock2018convnets, montavon2018methods, kim2019learning, xiao2020noise, teney2022evading, anders2022finding}. \emph{Explainable artificial intelligence} (XAI), a closely related area, analyzes the behavior of complex models, including deep neural networks, to better understand machine learning models~\cite{samek2019explainable}. Researchers have analyzed bias by using structured deep neural networks to visualize and interpret biases in data or models~\cite{phillips2009sample, mclaughlin2015data, tian2018eliminating, gunning2017explainable, adadi2018peeking, arrieta2020explainable}. Some approaches focus on estimating and quantifying bias levels in models and datasets to establish objective criteria for comparison~\cite{karkkainen2021fairface, zhao2021understanding}. To mitigate shortcuts, domain shift mitigation and data pruning methods have been proposed~\cite{kim2019learning, clark2019don, bahng2020learning, karkkainen2021fairface, nam2020learning}. Although various studies have proposed methods to mitigate dataset bias, such as data augmentation and data filtering~\cite{cohen2022study, geirhos2018imagenet, karkkainen2021fairface} as well as model-based approaches using bias predictors, adversarial training, and ensembles~\cite{kim2019learning, clark2019don, cadene2019rubi}, a systematic framework for identifying biases within a dataset remains lacking.

In this study, we propose a novel framework aimed at discovering shortcut learning in binary classifiers (detectors) \textcolor{black}{with existing datasets}. Given domain-specific information about the dataset, our framework enables the analysis of arbitrary \textbf{black-box} models, encompassing both shallow and deep architectures. In Section~\ref{BasicDefinitions}, we start by clearly defining key terminologies, as they are used variably across different works, and discussing related research. The proposed framework, outlined in Section~\ref{Sec:method} and detailed in Section~\ref{Section:MethodologyIntervs}, 
examines the impact of \textbf{both training and test data} on the classifier's performance. We approach the problem from both \textbf{interventional} 
and \textbf{observational} 
perspectives~\cite{thiese2014observational}. The former introduces controlled random interventions to training/test data to provoke reliance on dataset-level shortcuts, while the latter assesses the impact of observed or estimated nuisance features on classifier outputs. In both cases, our primary tool for modeling the dependency of biased data on classifier outputs is the \emph{linear mixed effects model} (LME)~\cite{lme4}, which enables direct regression of classifier detection scores to `go beyond reliance on error rates'.

\textcolor{black}{Focusing on the common element in all machine learning tasks—data—this work investigates whether potential shortcuts influence the learning process and, if so, to what degree. A key consideration, as discussed in this study, is the extent to which undesirable shortcut cues covary with the target (class) label. Additionally, our approach addresses both biased conditions and familiar non-biased scenarios, such as domain mismatch.}

Since our approach treats the analyzed model as a black-box score generator, in principle it is both model- and application-agnostic. Nonetheless, it is necessary to provide practical case studies to help link the theory and the practice. We consider two important \emph{speech information security} related tasks of \textbf{anti-spoofing} (real-fake discrimination---Section \ref{sec:antispoofing-case-study}) and \textbf{automatic speaker verification} (same-different speaker discrimination---Section \ref{sec:asv-case-study}). Besides their applications in call center security and audio forensics, both are foreseen as potential countermeasures against informed attacks (e.g. deepfakes and adversarial attacks) in social media and news platforms. Unfortunately, especially anti-spoofing models have been reported to be very sensitive to shortcuts (particularly the presence of silence)
\cite{muller21_asvspoof,chettri2018analysing,chettri2023clever,lapidot2019effects,liu2023asvspoof}. While somewhat less commonly said, speaker verification can also be subject to biases related to the data conditions. \textbf{The key contribution of our study a rigorous and unified formulation for bias analysis of detection models}, demonstrated here on two speech tasks.

Building on our preliminary study~\cite{Hyejin2023-coin-flip}, which explored interventions in audio anti-spoofing, this work significantly extends the methodology by incorporating a complementary observational perspective, introducing new anti-spoofing interventions with partial perturbations, adding a new detection task (speaker verification), and addressing \emph{partial} bias. Additionally, this study offers a more rigorous definition of models and their interpretations, while aiming to encourage reflection on the (over)reliance on error rates in experimental machine learning research, as discussed in Section~\ref{Section:Discussions}.

%% file: Section_Ib.tex
\section{Basic Definitions and Related Works}
\label{BasicDefinitions}

\textcolor{black}{In this section, we establish a clear foundation for understanding the key concepts relevant to this work, which are often used inconsistently across studies. By clarifying these definitions, we aim to promote a unified understanding and set the stage for discussing related work and methodologies.}

\subsection{Basic Definitions}

\textcolor{black}{\emph{Shortcut learning}: In machine learning, shortcut learning refers to a model that appears to make accurate predictions but relies on unintended patterns in the data rather than genuinely understanding the underlying task. This phenomenon is also known as the \emph{Clever Hans effect} in psychology~\cite{pfungst1911clever}, a \emph{confounder} in causality~\cite{pearl2009causality}, and a \emph{spurious correlation} in statistics~\cite{haig2003spurious} literature.}

\textcolor{black}{\emph{Bias}: In machine learning, "bias" has multiple meanings. It may refer to spurious correlations~\cite{ye2024spurious}, fairness concerns~\cite{mehrabi2021survey}, the trade-off between underfitting and overfitting~\cite{bishop2006pattern}, or systematic errors introduced by simplifying assumptions in a model~\cite{hastie2009elements,shah2020pitfalls}. Our study specifically addresses bias caused by spurious correlations, where unintended associations between covariates and target labels arise, often due to flawed data collection or experimental protocol. This is distinct from inclusiveness-related bias, which focuses on ensuring fair and equitable treatment across demographic groups (e.g., fairness concerning race, gender, or age)~\cite{barocas-hardt-narayanan2019-fairml}. While inclusiveness bias pertains to the ethical and social fairness of models, spurious correlation bias affects the technical reliability and generalization of their predictions~\cite{hutiri24_interspeech}. Our study specifically addresses biases arising from data-related artifacts, such as noise and compression, in anti-spoofing, as well as trial selection biases in speaker verification tasks. We refer to these biases collectively as \emph{dataset bias}.}

\textcolor{black}{\emph{Biased dataset}: The biased dataset refers to systematic issues in a dataset, such as imbalances, flawed sampling, or the presence of unintended patterns, that can lead machine learning models to make unreliable or unfair predictions by over-relying on spurious correlations rather than meaningful relationships~\cite{torralba2011unbiased,liu2024decade,moreno2012unifying}.}

\textcolor{black}{\emph{Biased model}: A biased model refers to a machine learning model that systematically produces skewed or unfair predictions due to inherent biases in its training process. These biases may arise from over-reliance on dataset-specific artifacts rather than learning generalizable patterns. As a result, such models may exhibit poor generalization across diverse scenarios, leading to unfair or inaccurate decisions~\cite{geirhos2020shortcut}.}

\textcolor{black}{\emph{Interventional and observational studies}: The observational approach involves examining relationships between variables without manipulation, simply by analyzing real-world data as it naturally occurs~\cite{parker2016planning}. In contrast, the interventional (or experimental) approach actively manipulates one or more variables in a controlled setting to assess their effects on other variables~\cite{thiese2014observational}. These complementary approaches offer valuable insights, with observational studies capturing natural data patterns and interventional studies enabling causal inference through controlled experimentation.}

\subsection{Related Studies}

\textcolor{black}{To mitigate unfair and unintended outcomes caused by dataset bias, understanding how models function and what information they learn has recently become a crucial research topic~\cite{torralba2011unbiased, stock2018convnets, montavon2018methods, kim2019learning, xiao2020noise, teney2022evading, anders2022finding}. Recent studies using various approaches have identified \emph{training data distribution} as a key source of bias~\cite{clark2019don, smilkov2017smoothgrad, karkkainen2021fairface, zhao2021understanding, bajorek2019voice, koenecke2020racial, phillips2009sample, mclaughlin2015data, tian2018eliminating}. For example, in the computer vision domain, biases may arise from the presence of specific keywords in captions or dataset-specific noise patterns~\cite{clark2019don, smilkov2017smoothgrad}. Similarly, models may focus on background elements rather than the target object, leading to unintended biases~\cite{phillips2009sample, mclaughlin2015data, tian2018eliminating}. Moreover, commonly occurring attributes contributing to bias often result in \emph{fairness} concerns, particularly regarding race, gender, and age~\cite{barocas-hardt-narayanan2019-fairml}. These disparities frequently stem from data imbalances across different subgroups, leading to unequal representation and biased outcomes~\cite{karkkainen2021fairface, zhao2021understanding, bajorek2019voice, koenecke2020racial}.}

\textcolor{black}{In addition to the aforementioned works in computer vision, several studies have also investigated dataset bias studies in the context of audio and speech related tasks as well. The classical study in~\cite{Sturm2014-Horse} demonstrates that a music information retrieval system may rely on dataset-specific artifacts rather than learning generalizable music features.
The study in~\cite{liu24f_interspeech} highlights the presence of shortcuts in speech-based Alzheimer's detection, where background noises captured in non-speech regions influence model predictions. Another recent study~\cite{gauder2024unreliability} also emphasizes the importance of using carefully curated datasets to produce reliable results in the Alzheimer's detection problem. The work in~\cite{oglic2022towards} developed a method to address common sources of spurious correlations in acoustic models for automatic speech recognition task.}

\textcolor{black}{Dataset bias is also evident in the two speech tasks examined in this study. In anti-spoofing, multiple studies have shown biases in publicly available speech corpora. For example, waveform sample distributions have been identified as shortcuts for spoofing detection~\cite{chettri2018deeper, chettri2018analysing, lapidot2019effects}. Additionally, differences in silence proportions between bonafide and spoofed data within a dataset can be exploited as learning shortcuts~\cite{zhang122021effect, muller21_asvspoof}. Such biases can affect detector training, hindering generalization across datasets~\cite{chettri21_interspeech}. Trimming silences~\cite{muller21_asvspoof} can mitigate silence-related shortcut learning by forcing classifiers to rely on active speech cues. Additionally, studies on ASVspoof 2021~\cite{liu2023asvspoof} explore data augmentation to enhance generalization for models trained on ASVspoof 2019 LA~\cite{wang2020asvspoof}.}


\textcolor{black}{While studies on spurious correlation-related bias in speaker recognition are limited, several works analyze bias concerning fairness~\cite{hutiri24_interspeech,estevez2023study, hajavi2023study, chouchane2023fairness, hutiri2022bias,toussaint22_interspeech,peri2023study}. Research in~\cite{rusti2023benchmark,leschanowsky2024data} examines voice biometrics datasets, highlighting bias and fairness issues without quantitative performance evaluation. Guidelines for fair VoxCeleb evaluation are proposed in~\cite{toussaint22_interspeech}, emphasizing trial selection effects on performance~\cite{estevez2023study}. A new metric for gender- and nationality-related biases is introduced in~\cite{hutiri24_interspeech}. Beyond biometrics, dataset bias also affects voice anonymization—a speaker-related task—with subgroup performance disparities reported in~\cite{williams2024anonymizing}~\cite{leschanowsky2024voice}. Speaker categorization into sheep, goats, lambs, and wolves~\cite{doddington98_icslp} further illustrates anonymization difficulty variations.}

\textcolor{black}{Although most studies experimentally demonstrate the presence of dataset biases and propose mitigation strategies, a theoretical framework for identifying and comparing the relative impact of different spurious correlation biases remains largely underdeveloped~\cite{Halder2024,Sreekumar2023}. Our study addresses this gap by introducing a structured approach with a simple linear model in two voice biometrics-related tasks.}

%% file: SECTION_III_Overview_of_Proposed_Framework.tex
\section{An Overview of the Proposed Framework}\label{Sec:method}

\begin{figure*}
    \begin{center}
        \includegraphics[scale=0.5]{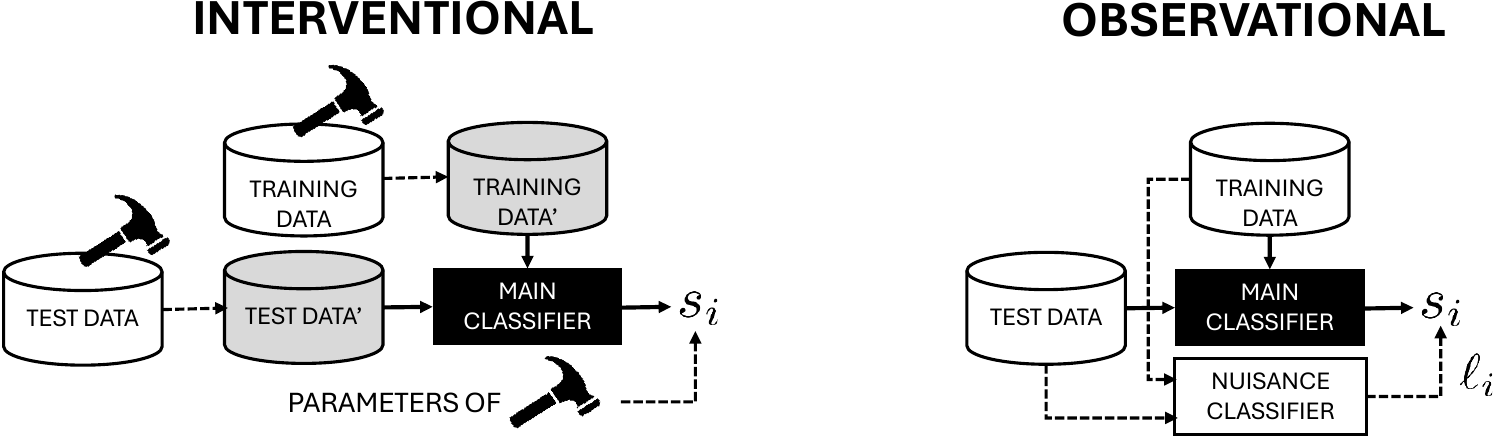}
        \caption{We model shortcut learning effect in binary classification tasks through two complementary approaches. The \textbf{interventional} method perturbs (e.g. adds noise to) an existing dataset to introduce systematic covariate shift to the class-conditional data distributions either on the training or test side. In the \textbf{observational} approach we assume the presence of systematic covariate shifts and extract relevant nuisance feature(s) $\ell_i$ (such as signal-to-noise ratio) that attempt to predict the class label based on the irrelevant features. In both approaches we model the dependency of the classifier output score $s_i$ (on either the intevention parameters or the nuisance score) using linear mixed effects (LME) modeling.}
        \label{fig:intervention-illustration} 
    \end{center}
    
\end{figure*}

Our proposed framework belongs to the class of \emph{explanatory} models of dataset biases in binary classification tasks. To be specific, we put forward a \emph{linear post-hoc} approach~\cite{rudin2022interpretable} that aims at quantifying dependence of the classifier score upon data: 
    \begin{mdframed}[style=MyFrame]
        \begin{center}
            Data $\rightarrow$ Classifier score
        \end{center}
    \end{mdframed}
where the arrow is to be read as `is a cause of'. This means that, keeping everything else (the model architecture, training algorithm, etc.) unchanged, variations in the data (either training and/or test) lead to different classifier outputs.

\textcolor{black}{We address the task from two complementary yet related perspectives: interventional and observational, both widely employed across various domains for causal inference and statistical modeling~\cite{thiese2014observational,li2023causal,zhang2022partial}.} The first, \emph{interventional} approach, as illustrated in Fig. \ref{fig:intervention-illustration} (and detailed in Section \ref{subsec:methodology-interventional}) introduces controlled random modifications to \textcolor{black}{an existing dataset}, so as to purposefully provoke the classifier to rely on shortcuts. 

The intervention process parameters are known, with  
other relevant variables  
fixed throughout each experiment, which facilitates  
causal analysis~\cite{castro2020causality}.  
In our second, \emph{observational} approach, as illustrated in Fig. \ref{fig:intervention-illustration} (and detailed in Section \ref{Section:MethodologyObs}), 
we extract suspected nuisance features(s) from data and include relevant measures of class disparity to our regression set-up (detailed below in Section~\ref{MixedEffectModel}).

In practice, our model-agnostic (black-box) approach quantifies the change in classifier outputs to changes in its inputs,  
through regression modeling of the score. This provides detail `beyond the error rate'.  
In the following, we briefly introduce the relevant background to binary classification and data mismatch that form the backbone for both the interventional and observational methodologies, covered in Section \ref{Section:MethodologyIntervs}. 

\subsection{The Common Pipeline of Fitting and Evaluating Detectors}

Let $\mathscr{D}\mydef \{(x_i,\vec{y}_i): i=1,\dots,N\}$ denote an evaluation dataset containing $N$ data instances, $x_i \in \mathscr{X}$, along with labels $\vec{y}_i \in \mathscr{Y}^2$ where $\mathscr{Y}\mydef\{0,1\}$. Each $\vec{y}_i$ is a 2-bit string $\vec{y}_i \mydef (y_i^\text{cls},y_i^\text{eva})$ where the first bit denotes the class (negative=0; positive=1) and the second bit denotes whether $x_i$ is placed to training or evaluation subset (train=0; evaluation=1). For instance, $\vec{y}_i=(1,0)$ indicates a positive class training instance. 
Conditioning by the subset (indicator) variable $\vec{y}_i$ yields partitioning of the data instances into four disjoint subsets, $X_{jk} \mydef \{x_i : \boldsymbol{y}_i=(j,k)\}$. For instance, $X_{00}$ and $X_{01}$ stand, respectively, for the training and test data of the negative class.

By using a suitable class of parametric predictive models $g_{\vec{\theta}}: \mathscr{X} \rightarrow \mathscr{Y}$ and training loss $\mathcal{L}: \mathscr{X}\times \mathscr{Y} \rightarrow \mathbb{R}_{+}$, a system developer trains a model (estimates a parameter vector $\vec{\theta} \in \mtx{\Theta}$) by minimizing $\mathcal{L}$ on the training data, $X_{00}\; \cup\; X_{10}$. The model is then executed on the evaluation data $X_{01}\;\cup X_{11}$ to produce predictions of class labels. In practice, the model produces a \emph{score} $s_i \in \mathbb{R}$ for each $x_i$, whereby larger relative values are associated with increased degree of membership of the positive class. The score can then be converted to a predicted binary label by comparing $s_i$ to a threshold $\tau$. The paired scores and the respective ground-truth labels $(s_i,y_i^\text{cls})$ are then used to estimate empirical miss and false alarm rates, each being a function of $\tau$. From the miss and false alarm rate functions, the evaluator finally summarizes performance using a suitable figure of merit (evaluation metric) applicable in one's research domain~\cite{krzanowski2009roc,brummer2006application,Hand2012-assessing-performance}. 

\subsection{Class Prior vs Data Distribution}\label{subsec:prior-vs-data-shift}

As noted above, we focus on modeling dependency of classifier upon the data it has been trained (and evaluated) on. To this end, \textcolor{black}{the dataset $\mathscr{D}$ can be viewed as a} random sample of paired instances $(x_i, y_i)$ from a fixed but unknown joint distribution $P(X,Y)$, 
which can be factorized as $P(X,Y)=P(X|Y)P(Y)$, 
where $P(Y)$ 
and $P(X|Y)$ 
denote, respectively, the class prior (prevalence) and the class-conditional data distributions. We introduce the notations 
    \begin{equation}
        \begin{aligned}
            \pi_\text{1$|$0} \mydef P(Y^\text{cls}=1|Y^\text{eva}=0)\\
           \pi_\text{1$|$1} \mydef P(Y^\text{cls}=1|Y^\text{eva}=1)\nonumber
        \end{aligned}
    \end{equation}
to denote, respectively, the positive class prior in the training and evaluation data. 
Further, we use
    \begin{equation}
        P_{jk}(X) \mydef P(X|\vec{Y}=(j,k))
    \end{equation}
to denote the respective sampling distributions of $X_{jk}$. 
With this notation, we can define different kinds of data-related mismatches \cite{Murphy2023-advanced-topics}:
    \begin{itemize}
        \item In \textbf{prior-matched} data, $\pi_\text{1$|$0}=\pi_\text{1$|$1}$ the class prevalence remains unchanged across training and test data, while $\pi_\text{1$|$0}\neq \pi_\text{1$|$1}$ corresponds to shift in the class prevalence. In \textbf{class-balanced} case, either training or test data corresponds to prior of $\frac{1}{2}$;
        \item In \textbf{covariate-matched} data, both $P_{00}(X)=P_{01}(X)$ and $P_{10}(x)=P_{11}(X)$ hold, i.e. training and test distribution for both classes are matched.
        \item When either $P_{00}(X)\neq P_{01}(X)$ or $P_{10}(X) \neq P_{11}(X)$ holds, we have \textbf{covariate mismatch}.
    \end{itemize}
In this study, we focus on systematic shift in the covariates, introduced through by interventions to the four conditional distributions $P_{ij}(X)$, detailed in Section~\ref{subsec:methodology-interventional}. 

\subsection{Mixed Effect Modeling of Biased Classifier Scores}
\label{MixedEffectModel}

\textcolor{black}{Whether due to purposeful controlled interventions applied to an existing data (detailed in Section \ref{subsec:methodology-interventional}), or due to latent nuisance factors present in it (detailed in Section \ref{Section:MethodologyObs}, we assume the classifier score $s_i$ to be dependent on some systematic, undesirable effects on the covariates (here, the audio data).}
\textcolor{black}{Even if the scoring function $x_i \mapsto s_i := g_{\vec{\theta}}(x_i)$ ---nowadays typically a deep neural network model---cannot be usually expressed using a simple analytic expression, we \emph{can} model the dependency of the detection score $s$ upon any relevant dependent variables available for regression modeling. This provides a simple, unified way to analyze the effects of training and/or test data to the classifier outputs.} 

\textcolor{black}{In practice, we use standard \emph{linear mixed-effects} (LME) model~\cite{lme4}, an extended family of conventional linear regression models that contains a mix of systematic and random effects. To address dataset biases\textemdash systematic disparities in data distributions\textemdash LME models are used in both interventional and observational settings. The generic form of LME models 
used in our study can be expressed as:}
\begin{equation}\label{eq:gen-LME}
     s_{i} = \mu + d y_i^\text{cls} + \vec{\beta}^\top \vec{u}_{i} + b_{i} + \varepsilon_{i},
\end{equation}
where:

\vspace{1ex}
    \begin{description}
        \item[$s_i$] the classifier score for trial $i$
        \item[$\mu$] the 
        mean score of the negative class
        \item[$d$] class discrimination (a fixed effect)
        \item[$y_i^\text{cls}$] the class label (0 for negative, 1 for positive)
        \item[$\vec{\beta}^\top\vec{u}_{i}$]the influence of intervention or nuisance variables (fixed effect)
        \item[${b}_i$] random effects accounting for trial-specific variability (e.g., speaker or recording conditions)
        \item[$\varepsilon_{i}$] the residual error accounting for other variabilities.
    \end{description}
\vspace{1ex}

\textcolor{black}{This model provides a unified framework for analysis. For the interventional case, $\vec{u}_{i}$ contains information on the controlled modifications applied to training or test data. In the observational case, it represents measured nuisance features (e.g., signal-to-noise ratio or speaker gender) that might influence classifier scores. By connecting fixed and random effects, LME enables a nuanced analysis of how dataset biases affect both positive and negative class distributions. This approach is crucial in tasks like speech anti-spoofing and speaker verification, where features can be associated to system's outcomes in a model that includes the effect of repeated group variability. }

\textcolor{black}{One may rightfully ask whether it is appropriate to use \emph{linear} models to explain scores that originate generally from highly \emph{nonlinear} scoring functions $g_{\vec{\theta}}$? Several neural network-based approaches, such as LMMNN~\cite{simchoni2023integrating}, MeNets~\cite{xiong2019mixed}, and DeepGLMM~\cite{tran2020bayesian}, have been developed as extensions of the linear model, allowing for the modeling of complex non-linear relationships between predictors and outputs. However, our aim here is to provide transparency to the impact of data-related quality factors upon classifier performance. Since \emph{explanations} are targeted for \emph{humans} \cite{Miller2019-explanation-in-AI,rudin2022interpretable}, there is a trade-off between model parsimony vs. suitability of explanations to our limited cognitive system. Hence, we deem it acceptable to use a linear approach to explain detection scores. After all, the scores (along with their class labels) are a complete description needed to derive any of the common performance measures (such as AUC, F1, and EER). Nonetheless, in the experiments we do also report the relevant criteria (adjusted $R^2$ statistic) to assess the goodness of fit of the linear models.}

%% file: SECTION_METHODLOGY.tex
\section{Methodology for Interventional and Observational Case}
\label{Section:MethodologyIntervs}

\subsection{Methodology: interventional case}
\label{subsec:methodology-interventional}

As illustrated in Fig. \ref{fig:intervention-illustration}, the key principle in our interventional approach is to anchor our experiments around an \emph{existing} dataset $\mathcal{D}$---for instance, a standard dataset used for reporting results in one's field. We then introduce specific random modifications to obtain a modified dataset, $\mathscr{D}'$
    \begin{mdframed}[style=MyFrame]
        \begin{center}
            $\mathscr{D}'$ = \text{Modify}($\mathscr{D}, \vec{\Lambda})$ 
        \end{center}
    \end{mdframed}
where `$\text{Modify}$' refers to an interventional procedure that returns a modified dataset according to a known statistical rule (i.e. fixed control parameters, $\vec{\Lambda}$). Selective modifications of $\mathscr{D}$, while retaining all other relevant aspects in an experiment unchanged, allows drawing conclusions on causal impact of data to the classifier score. The relevant considerations are:
    \begin{itemize}
        \item The total size of the intervened data remains unchanged,  $|\mathscr{D}|=|\mathscr{D'}|$;
        \item The labels $\vec{y}_i$ remain unchanged. Hence, the class prior and sizes of training/test data remain unchanged; 
        \item For each black-box, the architecture, training loss, numerical optimizer, number of training epochs, minibatch size, model selection criterion, and data augmentation recipes remain unchanged;
        \item The same computer environment is used for training and scoring each classifier. The experiment for a given black-box is conducted by the same experimenter. 
    \end{itemize}
What varies are the class-conditional data distributions $P_{ij}(X)$ denoted above.
As indicated in Fig. \ref{fig:intervention-illustration}, our intervention model is specified by three sets of parameters:
    \begin{itemize}
        \item The \textbf{type} of an intervention (additive noise, blanking, etc), implemented using a parametric deterministic function 
         $f(x; z)$, where $z$ denotes a control parameter.
        \item \textbf{Distribution} $P(z|\vec{\theta}_z)$ for the control parameter.
        \item Class-conditional \textbf{intervention probability}, denoted by $\rho_{f|\vec{y}} \mydef \mathbb{P}\big(\text{Apply } f \text{ to } x|Y=\vec{y})$;
    \end{itemize}
To be specific, a data instance $x_i$ in a subset specified by $\vec{y}$ gets randomly modified as 
    \begin{equation}\label{eq:interventional-algorithm}
        \begin{aligned}
        z_i & \sim_\text{i.i.d.} P(z|\vec{\theta}_z)\\
        x_ i' & = f(x_i; z_i),\;\;\; \text{with probability}\; \rho_{f|\vec{y}}.
        \end{aligned}
    \end{equation}
The intervention probability parameter controls \emph{how many of the data instances in a given subset get intervened}. For a subset of $M$ files, we apply $f$ to randomly selected $M_\text{interv} =\lfloor \rho_{f|\vec{y}}\cdot M\rfloor$ files, the remaining $M - M_\text{interv}$ being retained as in the original dataset. The control parameter $z_i$, in turn, is a univariate control parameter. As an example, in our experiments with white additive noise, $z_i$ is a randomly signal-to-noise ratio (SNR) drawn from a uniform distribution. 





How do we choose the interventions, their control parameter distributions and the intervention probabilities? For tractability and avoidance of combinatorial explosion in our experiments, we make a number of simplifications. As in \cite{Hyejin2023-coin-flip}, we focus on each type of intervention at a time: we neither mix different interventions types nor consider their combinations. Second, $P(z;\vec{\theta}_z)$ takes the form of a continuous uniform distribution. 
Besides stochastic interventions, our experiments also include $\mu$-law~\cite{kondoz2005digital} encoding, a parameter-free deterministic operation. 

\text{black}{Table~\ref{tab:config} presents the various possible intervention conditions, where one or both classes (positive and negative) in the training and test subsets are modified.} Our main interest lies in the four intervention probabilities $\rho_{f|\vec{y}}$, also referred to as \emph{configurations} in Table \ref{tab:config}. In these \emph{corner cases}, the four probabilities are constrained to be binary ($\rho_{f|\vec{y}} \in \{0,1\}$), i.e. perturbing either \emph{all} or \emph{none} of the files in the specified data subset. While in \cite{Hyejin2023-coin-flip} only these corner cases were considered, here we consider the generalized \emph{partially biased} set-up with arbitrary specified intervention proportions for the positive and negative classes. Due to computational reasons, we limit the training side interventions to the corner cases, but let the test side probabilities to vary over a grid.

For the regression models, it is convenient to parameterize the intervention process in terms of two independent variables, $\vec{u}_i \equiv (\Delta_{i}^\text{1}, \Delta_{i}^\text{0})^\top$. The first one, $\Delta_{i}^\text{1}$, is defined as the absolute difference between the intervention probability of the test instance $i$ and the intervention probability of the positive class training set. Likewise, $\Delta_{i}^\text{0}$ is defined the absolute difference between the intervention probability of the test trial and the intervention probability of the spoof training set. The value $0$ indicates an equivalent treatment of test and training audio, while 1 indicates different treatments. 

\subsection{Methodology: observational case}
\label{Section:MethodologyObs}

We now proceed to the case of modeling \emph{observational} data. 
The difference to the interventional case is that 
now the dataset is given \emph{as-is}. 
Nonetheless, similar to the $z_i$ variables in \eqref{eq:interventional-algorithm}, we assume the existence of 
variable(s) associated with each audio file that may influence the classifier score. These variables can be observed (given) or unobserved (latent), the latter requiring values estimated from the provided audio data. Both cases can be handled in a unified way 
by including the given (or measured) variables 
as independent parameters in the LME model. 

\textbf{Observed (given) nuisance variables:} Familiar examples are 
categorical \emph{condition variables} beyond the class label---for instance, speaker's gender/language/age/nationality; or a particular recordings's location, device, or codec details. When available, such \emph{metadata} variables allow one to break-down performance across various subsets of the data and monitor for potential discrepancies across demographic or other kinds of performance factors. We defer to a detailed discussion in the context of speaker verification until Section \ref{sec:asv-case-study}.

\textbf{Unobserved (latent) nuisance variables:} Examples 
are speech datasets recorded under additive background noise or reverberant room, but with uncertain knowledge of the ground-truth noise type, SNR, reverberation time, etc. In the following, we detail our practical approach to estimate the relevant nuisance variables from speech data. \textcolor{black}{Our premise is that, similar to given metadata or condition variables, the measured variables should be explainable to the database designer, in order to help revise either the audio data and/or the training-test protocol.} 




In our approach, the database designer estimates any relevant \emph{nuisance feature} $w_i$, such as SNR\footnote{\url{http://labrosa.ee.columbia.edu/projects/snreval/}} or proportion of nonspeech, from each data instance $x_i$.
In this work, we assume $w_i$ to be a scalar. We use the nuisance features extracted from the training data of the two classes to train a \emph{nuisance classifier} that aims to predict the class label. 
Ideally, we prefer this classifier to \emph{fail}, assuming $w_i$ is a feature we prefer our main classifiers to not rely on. 
For instance, in speech anti-spoofing and speaker verification it might be undesirable to rely on channel, noise or silence cues. 

In practice, the nuisance classifier computes a \emph{log-likelihood ratio} (LLR) score between the two hypotheses:
    \begin{equation}\label{eq:perturbation-LLR-computation}
        \ell_i \mydef \log\frac{p\big(w_i|\vec{Y}=(1,0))}{p\big(w_i|\vec{Y}=(0,0)\big)}\approx \log\frac{p\big(w_i|\vec{\theta}_\text{10}\big)}{p\big(w_i|\vec{\theta}_\text{00})},
    \end{equation}
where `$\approx$' signifies the use of parametric statistical models in place of the true (unknown) likelihood functions. Here, $\vec{\theta}_\text{10}$ and $\vec{\theta}_\text{00}$ denote class-conditional parameters obtained from the training set. 
Since the nuisance features $w_i$ are univariate, with 
thousands of training instances, 
we fit a \emph{Gaussian mixture models} (GMM) separately to the training data of each class using the \emph{expectation-maximization} (EM) algorithm \cite{dempster1977}. 

\textcolor{black}{Similar to the `$\Delta_i$' variables in our interventional model, the $\ell_i$ 
variables quantify potential bias towards one of the classes. From a database designer's perspective, the ideal value is $\ell_i=0$ for all test trials. Values with large magnitude $|\ell_i| \gg 0$ indicates potential bias. In our LME modeling, we simply treat $\ell_i$ as another independent variable to gauge its effect to the dependent variable, i.e. the detection score $s_i$. Note that \eqref{eq:perturbation-LLR-computation} gauges bias in one specific test instance $x_i$. From the measurements $\{\ell_i\}$ gathered across all the test instances we can obtain distribution-level bias measures, useful for assessing the overall bias in the training-test protocol. We point the interested reader to Appendix \ref{appendix:bias-score-interpretation} for this discussion. Under certain assumptions, these measures are not dependent on the choice of the nuisance classifier in \eqref{eq:perturbation-LLR-computation}.}

%% file: SECTION_VI_Application_to_Antispoofing.tex
\section{Case study I: Anti-Spoofing}\label{sec:antispoofing-case-study}

We first consider the task of \emph{anti-spoofing}, which involves labeling a speech utterance as either 'bonafide' (real human sample) or 'spoofing attack' (machine-generated).



\subsection{Experimental Setup}

We use the ASVspoof 2019 logical access (LA) dataset~\cite{wang2020asvspoof} for all the anti-spoofing experiments. It consists of speech synthesis and voice conversion samples distributed across training, development, and evaluation subsets. The two former include six types of spoofing attacks, while the last contains thirteen attacks.



While prior studies have focused mainly on silence-related biases, our study provides substantially expanded set of interventions: MP3 compression, additive white noise, loudness normalization, non-speech blanking, and $\mu$-law compression. The interested reader is pointed to our preliminary study~\cite{Hyejin2023-coin-flip} for further detail.

\input{tables/corner_config}
\input{tables/models_config}

To demonstrate agnosticity of our methodology to the classifier under analysis, we consider two entirely different anti-spoofing 'black-boxes', one representative of hand-crafted features and statistical classifier, another based on modern deep learning based model. The former uses linear frequency cepstral coefficient (LFCC) features with Gaussian mixture model (GMM)~\cite{sahidullah15_interspeech}. The latter is
AASIST~\cite{jung2022aasist}, a state-of-the-art graph neural network based approach that operates upon raw waveforms. 


For the analysis of detection scores in the interventional case, all scores are normalized with Z-score separately for each configuration and intervention. The regression coefficients are estimated using \textit{lme4 package}~\cite{lme4} for \texttt{R}.

To analyze the audio anti-spoofing model using proposed method, we conducted largely four experiments: (i) Interventional Case, (ii) Partially Interventional Case, (iii) Observational Case.

Table~\ref{tab:config} presents the various dataset intervention conditions adopted for the anti-spoofing evaluation. For observational studies, we utilize the original data, which is also included in the same table as \textbf{O}.

\input{tables/cm_results}

\subsection{Results: Interventional Case}\label{Section:Results}


Table~\ref{tab:cm_performance} shows the comparative performance of the two countermeasure models for various interventions as explained in Table~\ref{tab:config}. 
In the configuration perspective, the configurations with intervention in the same class across training and test (i.e, \(\textbf{IT}_{n}\) and \(\textbf{IT}_{p}\)) substantially reduce EER compared to other configurations. 
For some cases, they show extreme case performance with 0\% EER indicating perfect discrimination.
On the other hand, completely opposite trends are shown when the intervention is reversed for classes across training and test (i.e., \(\textbf{IV}_{pn}\) and \(\textbf{IV}_{np}\)). In some cases, we observe more than 50\% EER indicating label flipping. 
Those results support the dataset bias acts as an additional cue for discrimination. 

In terms of the type of interventions, the models are highly sensitive, particularly in MP3 compression and additive white noise. The less significant intervention involves loudness normalization through countermeasure models. An important observation is that when intervention is added to spoof utterances in training, it has a greater impact compared to adding intervention to bona fide utterances in training on neural classifiers, shown in the gap between \(\textbf{IV}_{pn}\) and \(\textbf{IV}_{np}\). We can reasonably conclude that bona fide speech is less susceptible to silence, which aligns with the findings of~\cite{muller21_asvspoof}. Additionally, the trend across different features and models is consistent, indicating a potential dataset bias.

\input{tables/mixed_effects}

\subsection{Partial Bias}

\begin{figure}[!t]
    \centering

    \subfloat[\footnotesize{Original training data for both classes.}]{
        \includegraphics[
            trim={0 5cm 0 0.5cm},
            clip,
            width=1.1\linewidth
        ]{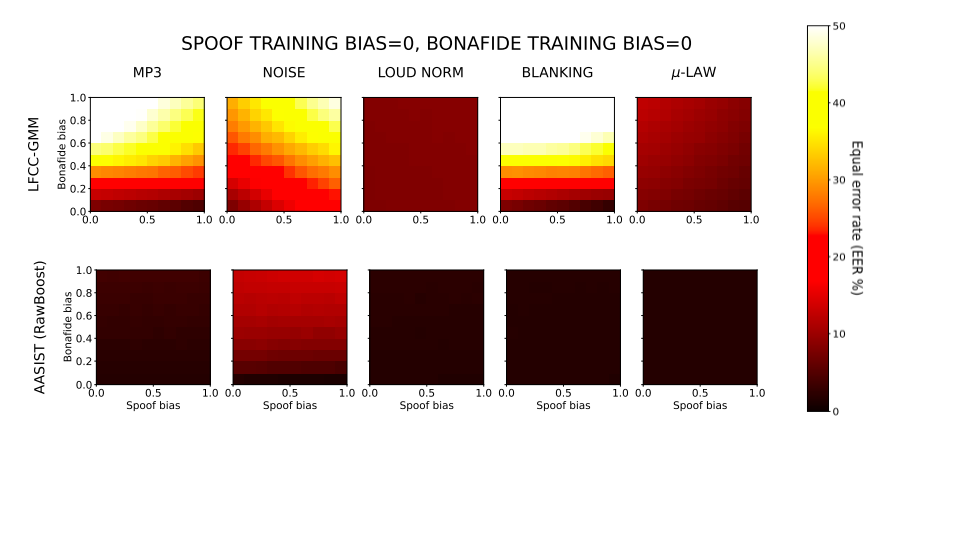}
    }

    \subfloat[\footnotesize{Original training data for spoof, biased training data for bona fide.}]{
        \includegraphics[
            trim={0 5cm 0 0.5cm},
            clip,
            width=1.1\linewidth
        ]{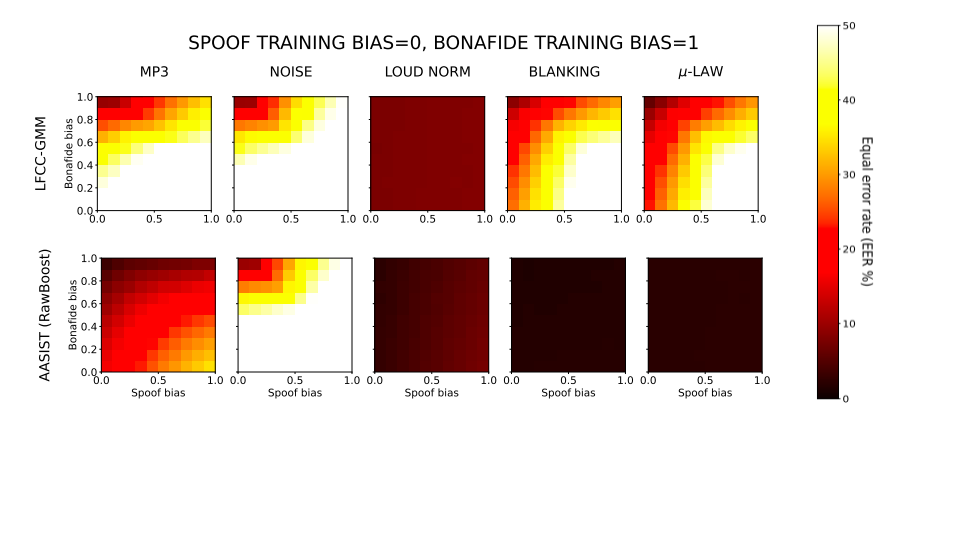}
    }

    \subfloat[\footnotesize{Original training data for bona fide, biased training data for spoof.}]{
        \includegraphics[
            trim={0 5cm 0 0.1cm},
            clip,
            width=1.1\linewidth
        ]{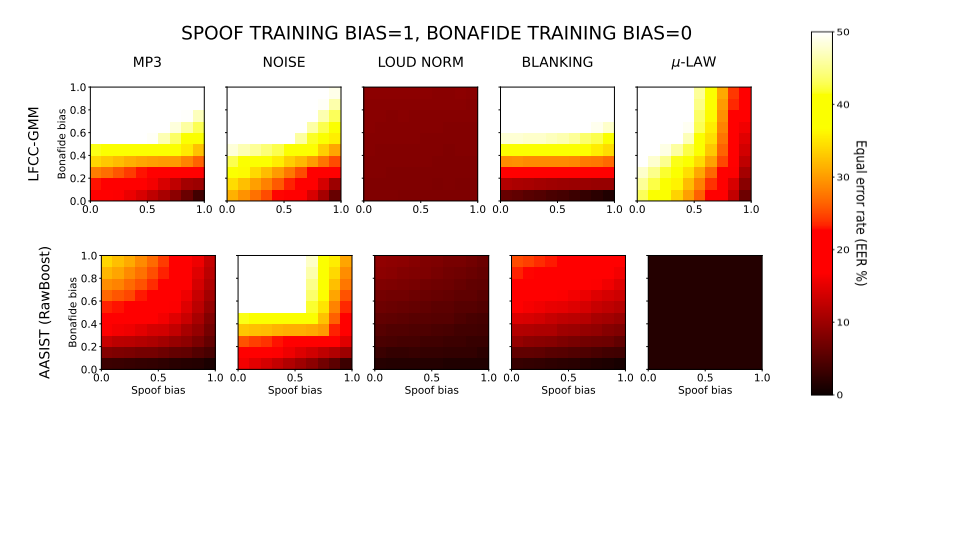}
    }

    \vspace{-0.2cm}
    \caption{Heatmaps of bonafide-spoof EER for five different interventions and two spoofing countermeasures (LFCC-GMM~\cite{sahidullah15_interspeech}, AASIST with RawBoost~\cite{tak2022rawboost}). The three sets of panels correspond to different training data interventions (a: no interventions; b: intervention to bona fide class only; c: intervention to spoof training only), and each of the fifteen panels displays the results for varied degrees of test-side intervention probability.}
    \label{fig:partial-bias-graphs}
    \vspace{-0.5cm}
\end{figure}

As a generalization of the four special cases I, M, IT, and IV, consider now the results of \emph{partially} biased setups visualized in Fig. \ref{fig:partial-bias-graphs}. 
The four corners of each of the panels in Fig. \ref{fig:partial-bias-graphs} are obtained by letting $\vec{\rho} \mydef (\rho_0, \rho_1)$ to take values $(0,0)$, $(0,1)$, $(1,0)$ or $(1,1)$. Depending on the training condition, the interpretation of these corner values is different. 
X and Y axes refer to the level of bias in test data for spoof and bonafide class, respectively.  

From the top, Fig. \ref{fig:partial-bias-graphs} (a) with original training data for both classes shows that the overall performance is not affected a lot compared to other training condition. 
However, in the case of MP3, NOISE, and BLANKING, the performance degrades gradually impacted by intervention on the bona fide side and is shown as horizontal gradation.
It can be interpreted as bona fide data is sensitive or related to those interventions. 
This phenomenon is most distinct in LFCC-GMM system.

On the other hand, Fig. \ref{fig:partial-bias-graphs} (b) and (c), which include bias in either bonafide or spoof class in the training, show different trend compared with Fig. \ref{fig:partial-bias-graphs} (a). 
Both show prominent opposite diagonal correlation in most intervention cases.
In particular, those relation is noticeable in NOISE with all systems. 
AASIST with RawBoost can be seen as the most robust, but it is also largely affected by NOISE.



\subsection{Modeling of Biased CM Scores}\label{Sec:MixeEffectModel}

Although EER provides insights into the dependence of a countermeasure on training and test data interventions, it lacks granularity on class-specific score variations. For instance, an increased EER between intervention configurations in Table \ref{tab:config} does not indicate whether it stems from higher negative class scores, lower positive class scores, or both. Direct regression modeling of CM detection scores using interventions as independent variables allows a deeper analysis beyond `go beyond the EER'. Assume a CM is trained and scored on dataset $\mathscr{D}$ treated with a configuration from Table \ref{tab:config}.

With full knowledge of intervention parameters, they serve as independent variables to model the detection scores' statistical dependency. Thus, we employ linear modeling of biased CM scores. For the $i$th test trial, the specific model derived from the generic model in \eqref{eq:gen-LME} is given by

    \begin{equation}\label{eq:llr-mixed-effects-model}
     s_{i} = \mu + d\, y_i^\text{cls} + \beta^{\text{bon}}\Delta_{i}^\text{bon} + \beta^{\text{spf}}\Delta_{i}^\text{spf}
     + \varepsilon_{i},
    \end{equation}

In this case, $\varepsilon_{i}$ absorbs \emph{random effects} related to between-trial variation.

A concrete example may be helpful in clarifying \eqref{eq:llr-mixed-effects-model}. For the configuration $\textbf{IT}_{b}=\texttt{(0101)}$ in Table \ref{tab:config}, $\Delta_i^\text{bon}=0$ and $\Delta_i^\text{spf}=1$ for all the bona fide trials; and $\Delta_i^\text{bon}=1$ and $\Delta_i^\text{spf}=0$ for all the spoof trials. 
The two class-conditional models obtained from \eqref{eq:llr-mixed-effects-model} are 
 \begin{equation}
    \begin{aligned}
        s_i & = \mu + \beta^{\text{bon}} + \varepsilon_{i} & \text{(spoof trials, $y_i^\text{cls}=0$)}\\
        s_i & = \mu + d + \beta^{\text{spf}} + \varepsilon_{i} & \text{(bona fide trials, $y_i^\text{cls}=1$)}
    \end{aligned}
  \end{equation}
Since $\varepsilon_i$ is normal, both of these conditional score distributions are normal as well, with shared variance $\sigma_\varepsilon^2$. The difference between the bona fide and spoof class means (which relates to discrimination) is $d + (\beta^\text{spf}-\beta^\text{bon})$. The expression in the parentheses vanishes when the two classes are treated the same way (original configuration \textbf{O}). Whenever the difference of $\beta^\text{spf}$ and $\beta^\text{bon}$ is positive, the separation of the two distributions improves, leading to `decreased' EER relative to  \textbf{O}. Likewise, a negative difference yields an `increase' in EER. The use of quotes is intentional, as the $\beta$-coefficients relate to the systematic effects that we introduced to the data through the interventions. 

Similar model interpretations are easy to obtain for the remaining configurations; see Table \ref{tab:config_models}. As one might expect, the model for the two domain intensification configurations ($\textbf{IT}_{p}$ and $\textbf{IT}_{n}$) are the same; likewise, the model for the two domain inversion (label-flip) configurations ($\textbf{IV}_{pn}$ and $\textbf{IV}_{np}$) are the same. Considering the difference of the class-conditional means, the only difference between the two sets of biased models is in the sign of $(\beta^\text{spf}-\beta^\text{bon})$ or $(\beta^\text{bon}-\beta^\text{spf})$.
In our result analysis, we use $\beta^*$ referring to $\beta^{\text{bon}}$ or $\beta^{\text{spf}}$ which only differ in the sign with $\beta^{\text{spf}} = \beta^*$ and $\beta^{\text{bon}} =  - \beta^*$. 





We employed a mixed-effect model with Eq.~\eqref{eq:llr-mixed-effects-model} to fit the standardized detection scores and determine the effect of configuration variation on the bias across the trials. For each intervention, we consider a single model that corresponds to the five configurations (i.e., \textbf{O}, \textbf{IV}, \textbf{IT}). Table~\ref{tab:lm_models} shows the parameters of each CM model on each intervention. We found a substantial effect on $\beta^*$ (referring to $\beta^{\text{bon}}$ or $\beta^{\text{spf}}$) for all the models except for loudness normalization. In terms of interventions, MP3 compression, and additive white noise showed higher variation effects for $\beta^*$, while loudness normalization produced smaller variation effects for the data. 

\subsection{Linear Models for the Observational Data}

For pedagogical reasons, until this point, the authors have purposefully constrained the focus on simple linear regression approaches, refraining discussion on the random effects that have been thus far included to the residual term. Nohenthless, since the `salt' of the LME models is exactly in their ability to handle both fixed and random effects, the latter being clearly separated from the model fitting error (residual). Towards this, our LME model includes \emph{random effects}, which were previously represented as part of the residual term. Our data has group effects corresponding to audio from different speakers, and spoof audio produced by different types of spoofing attacks. These are considered as random effects that can be drawn from a population of speakers or types of attacks and are considered independent among groups. In this model, we can expect trial variability per-speaker represented as ${b}_i \sim \mathcal{N}(0,\sigma_b^2)$ and per-type of spoof attack as ${c}_i \sim \mathcal{N}(0,\sigma_c^2)$, where $\Sigma$ is the variance-covariance matrix for the random effects. Then besides the residual variance, $\sigma_\varepsilon^2$, we include random effects variance for the speaker effect, $\sigma_b^2$, and for the type of spoof attack $\sigma_c^2$ with ${b}_i$ and ${c}_i$ are added to equations \eqref{eq:llr-mixed-effects-model}. These variations will show adjustments per-speaker and per-attack type to the intercepts of the model. The updated LME model for equations~\eqref{eq:llr-mixed-effects-model} are:
\vspace{-0.15cm}
    \begin{equation}\label{eq:classconditional-lme-for-observational-data-with-random}
    \begin{dcases}
        s_i = \mu + \beta_\ell^\text{spf}\ell_i + b_i + c_i + \varepsilon_i, & \text{for } y_i^\text{cls} = 0 \\
        s_i = \mu + d + \beta_\ell^\text{bon}\ell_i + b_i + c_i + \varepsilon_i, & \text{for } y_i^\text{cls} = 1,
    \end{dcases}
    \end{equation}

\subsection{Results: Observational Case}\label{Section:ResultsObs}

\textcolor{black}{The results of the LME model~\eqref{eq:classconditional-lme-for-observational-data-with-random} for the original data (\textbf{O}) are presented in Table~\ref{tab:intervention_snr_blanking} for the GMM system and the AASIST with the RawBoost system. In addition to the model parameters, the table includes the \emph{conditional} $R^2$, a measure of the model’s goodness of fit that accounts for both fixed and random effects in the mixed-effects framework.}

\textcolor{black}{This observational analysis utilizes the log-likelihood ratio (LLR) score of the nuisance classifier as a predictor ($\ell_i$), where $\ell_i$ is computed using either the estimated noise level or the non-speech proportion of the $i$-th trial. These predictor variables, defined in~\eqref{eq:classconditional-lme-for-observational-data-with-random}, capture variations in the input conditions that may influence the model’s performance. The results shows a substantial class discrimination parameter ($d$) for both the estimated SNR and non-speech proportions with two different classifiers. The non-speech proportion shows higher values of $\beta$s indicating that it has a stronger influence on the model’s predictions compared to the estimated SNR. This suggests that variations in the proportion of non-speech content on the original data contribute more significantly to the observed class separation. The results from this LME analysis with both the classifiers and data-related factors also indicate that the random effect associated with variation in attacks (i.e., $\sigma_c^2$) is greater than that due to variation in speakers (i.e., $\sigma_b^2$). Moreover, these variations do not differ noticeably across the two speech data-related measures for each individual classifier.}

\input{tables/intervention_snr_blanking}

%% file: tables/corner_config.tex
\begin{table}[!t]
    \caption{\textcolor{black}{We consider \emph{biased} detection tasks where selected interventions are applied to one of the four subsets ($\{$negative, positive$\} \times \{$ train, test$\}$). Subscripts $p$ and $n$ in configurations refer to the class. The binary flags \texttt{0} and \texttt{1} indicate whether or not the specific subset undergoes interventions. From the top row, \textbf{O} and \textbf{I} are original and fully intervened cases, respectively. In \textbf{M} configurations, the treatment is applied to both classes but only either to train ($\textbf{M}_{\mathrm{tr}}$) or test data ($\textbf{M}_{\mathrm{te}}$), a form of simulated domain mismatch. In \textbf{IT} and \textbf{IV} cases, only one of the classes (in both train and test) undergoes intervention, leading to intensified (\textbf{IT}) or inverted (\textbf{IV}) association between data and class label. If intervention is applied to different classes during training and test, it is expressed as \(\textbf{IV}_{\mathrm{np}}\) and \(\textbf{IV}_{\mathrm{ps}}\), following the train and test in sequence. \textbf{O} with subscripts are special cases that are similar case with \textbf{O} on the training side, but only different on the test.}}
    \vspace{-0.2cm}
    \label{tab:config}
    \centering
    \begin{tabular}{c|l|c|}
    \hline
    &\multirow{2}{*}{\textbf{Configuration}} & \multirow{2}{*}{\textbf{Category}}  \\ 
    &\texttt{(indicator)}&     \\ \hline
    \parbox[t]{2mm}{\multirow{4}{*}{\rotatebox[origin=c]{90}{Fair eval}}}&\textbf{O}\texttt{(0 0 0 0)} & Original   \\ \cline{2-3}
    &\textbf{I}\texttt{(1 1 1 1)} & Totally Intervened \\ \cline{2-3} \cline{2-3}
    &$\textbf{M}_{\mathrm{tr}}$\texttt{(1 1 0 0)} & \multirow{2}{*}{Domain Mismatched}  \\ \cline{2-2}
    &$\textbf{M}_{\mathrm{te}}$\texttt{(0 0 1 1)} &    \\ \hline
    \hline
    
    \parbox[t]{2mm}{\multirow{4}{*}{\rotatebox[origin=c]{90}{Bias eval}}}&$\textbf{IT}_{\mathrm{p}}$\texttt{(0 1 0 1)} & \multirow{2}{*}{Domain InTensified}\\ \cline{2-2} 
    &$\textbf{IT}_{\mathrm{n}}$\texttt{(1 0 1 0)} &     \\  \cline{2-3}
    &$\textbf{IV}_{\mathrm{pn}}$\texttt{(0 1 1 0)} & \multirow{2}{*}{Domain InVerted} \\ \cline{2-2}
    &$\textbf{IV}_{\mathrm{np}}$\texttt{(1 0 0 1)} &    \\ \hline \hline
    \parbox[t]{2mm}{\multirow{2}{*}{\rotatebox[origin=c]{90}{Other}}}&$\textbf{O}_{\mathrm{n}}$\texttt{(0 0 1 0)} & Original training,  \\ \cline{2-2} 
    &$\textbf{O}_{\mathrm{p}}$\texttt{(0 0 0 1)} & one class intervention in test  \\ \cline{2-3}
    \hline
\end{tabular}
\end{table}

%% file: tables/models_config.tex
\begin{table}[tb]
\caption{Models for anti-spoofing observational data per configuration and trial class $y_i^\text{cls}$ (0: spoof, 1: bonafide), where \textit{Difference}  refers to the difference of conditional means: $\E[s_i|y_i^\text{cls}=1] - \E[s_i|y_i^\text{cls}=0]$}.
\label{tab:config_models}
\vspace{-5pt}
\resizebox{\columnwidth}{!}{%
\begin{tabular}{llll}
\hline
Config. & $y_i^\text{cls}$ &Model & \textit{Difference} \\ \hline
\textbf{O}, \textbf{I}    & 0 &$s_i  = \mu + \varepsilon_{i}$ & $d$ \\
              & 1 &$s_i  = \mu + d+ \varepsilon_{i}$ &  \\ \hline
$\textbf{M}_{\mathrm{tr}}$, $\textbf{M}_{\mathrm{te}}$  & 0 & $s_i  = \mu + \beta^{\text{bon}} + \beta^{\text{spf}} + \varepsilon_{i}$ & $d$ \\
            & 1 & $s_i = \mu + d + \beta^{\text{bon}} + \beta^{\text{spf}} + \varepsilon_{i}$ & \\ \hline
$\textbf{IT}_{p}$, $\textbf{IT}_{n}$ & 0 & $s_i  = \mu + \beta^{\text{bon}} + \varepsilon_{i}$ & $d +\beta^{\text{spf}} - \beta^{\text{bon}}$  \\
            & 1 & $s_i  = \mu + d + \beta^{\text{spf}} + \varepsilon_{i}$ &   \\ \hline
$\textbf{IV}_{\mathrm{pn}}$, $\textbf{IV}_{\mathrm{np}}$  & 0 & $s_i  = \mu + \beta^{\text{spf}} + \varepsilon_{i}$ & $d +\beta^{\text{bon}} - \beta^{\text{spf}}$\\
            & 1 & $s_i = \mu + d + \beta^{\text{bon}} + \varepsilon_{i}$ & \\ \hline
$\textbf{O}_{n}$, $\textbf{O}_{p}$  & 0 & $s_i  = \mu + \beta^{\text{bon}} + \beta^{\text{spf}} + \varepsilon_{i}$ & $d$ $ - \beta^{\text{bon}} - \beta^{\text{spf}}$ \\
            & 1 & $s_i = \mu + d +  \varepsilon_{i}$ & \\ 
            \hline
\end{tabular}%
\vspace{-0.5cm}
}
\end{table}

%% file: tables/cm_results.tex
\begin{table*}[]
\setlength{\tabcolsep}{20pt}
\renewcommand{\arraystretch}{1.2}
  \centering
  \scriptsize
  \caption{Countermeasures performance (in EER \%) in all configurations defined in Table~\ref{tab:config}.}
  \vspace{-0.15cm}
  \label{tab:cm_performance}
  \setlength{\tabcolsep}{0.5em}
\begin{tabular}{l|ccccccccc|ccccccccc|}
\cline{2-19}
 & \multicolumn{9}{c|}{\textbf{GMM}} & \multicolumn{9}{c|}{\textbf{AASIST+RawBoost}} \\ \hline
\multicolumn{1}{|l|}{Intervention} & \multicolumn{1}{c|}{\textbf{I}} & \multicolumn{1}{c|}{$\textbf{M}_{tr}$} & \multicolumn{1}{c|}{$\textbf{M}_{te}$} & \multicolumn{1}{c|}{$\textbf{IT}_{p}$} & \multicolumn{1}{c|}{$\textbf{IT}_{n}$} & \multicolumn{1}{c|}{$\textbf{IV}_{ps}$} & \multicolumn{1}{c|}{$\textbf{IV}_{np}$} & \multicolumn{1}{c|}{$\textbf{O}_{n}$} & \multicolumn{1}{c|}{$\textbf{O}_{p}$}  & \multicolumn{1}{c|}{$\textbf{I}$} & \multicolumn{1}{c|}{$\textbf{M}_{tr}$} & \multicolumn{1}{c|}{$\textbf{M}_{te}$} & \multicolumn{1}{c|}{$\textbf{IT}_{p}$} & \multicolumn{1}{c|}{$\textbf{IT}_{n}$} & \multicolumn{1}{c|}{$\textbf{IV}_{pn}$} & \multicolumn{1}{c|}{$\textbf{IV}_{np}$} & \multicolumn{1}{c|}{$\textbf{O}_{n}$} & \multicolumn{1}{c|}{$\textbf{O}_{p}$}  \\ \hline
\multicolumn{1}{|l|}{MP3} & \multicolumn{1}{c|}{12.23} & \multicolumn{1}{c|}{22.49} & \multicolumn{1}{c|}{44.83} & \multicolumn{1}{c|}{0.00} & \multicolumn{1}{c|}{0.00} & \multicolumn{1}{c|}{99.99} & \multicolumn{1}{c|}{97.85} & \multicolumn{1}{c|}{4.01} & \multicolumn{1}{c|}{61.26}  & \multicolumn{1}{c|}{3.12} & \multicolumn{1}{c|}{2.31} & \multicolumn{1}{c|}{3.71} & \multicolumn{1}{c|}{0.88} & \multicolumn{1}{c|}{0.52} & \multicolumn{1}{c|}{35.96} & \multicolumn{1}{c|}{36.03} & \multicolumn{1}{c|}{1.62} & \multicolumn{1}{c|}{3.96} \\ \hline
\multicolumn{1}{|l|}{Noise} & \multicolumn{1}{c|}{23.59} & \multicolumn{1}{c|}{43.51} & \multicolumn{1}{c|}{52.55} & \multicolumn{1}{c|}{0.00} & \multicolumn{1}{c|}{0.01} & \multicolumn{1}{c|}{99.98} & \multicolumn{1}{c|}{99.99} & \multicolumn{1}{c|}{21.06} & \multicolumn{1}{c|}{32.28}  & \multicolumn{1}{c|}{4.05} & \multicolumn{1}{c|}{3.50} & \multicolumn{1}{c|}{14.71} & \multicolumn{1}{c|}{0.22} & \multicolumn{1}{c|}{0.01} & \multicolumn{1}{c|}{99.82} & \multicolumn{1}{c|}{99.95} & \multicolumn{1}{c|}{1.06} & \multicolumn{1}{c|}{13.27}  \\ \hline
\multicolumn{1}{|l|}{Loudness} & \multicolumn{1}{c|}{8.22} & \multicolumn{1}{c|}{7.73} & \multicolumn{1}{c|}{8.70} & \multicolumn{1}{c|}{7.61} & \multicolumn{1}{c|}{7.83} & \multicolumn{1}{c|}{8.44} & \multicolumn{1}{c|}{9.00} & \multicolumn{1}{c|}{8.28} & \multicolumn{1}{c|}{8.32} & \multicolumn{1}{c|}{2.98} & \multicolumn{1}{c|}{2.31} & \multicolumn{1}{c|}{2.31} & \multicolumn{1}{c|}{2.20} & \multicolumn{1}{c|}{1.32} & \multicolumn{1}{c|}{7.79} & \multicolumn{1}{c|}{9.88} & \multicolumn{1}{c|}{1.52} & \multicolumn{1}{c|}{2.38}  \\ \hline
\multicolumn{1}{|l|}{Nonspeech} & \multicolumn{1}{c|}{31.79} & \multicolumn{1}{c|}{11.62} & \multicolumn{1}{c|}{65.10} & \multicolumn{1}{c|}{2.40} & \multicolumn{1}{c|}{0.57} & \multicolumn{1}{c|}{81.67} & \multicolumn{1}{c|}{90.53} & \multicolumn{1}{c|}{1.82} & \multicolumn{1}{c|}{87.18}  & \multicolumn{1}{c|}{1.62} & \multicolumn{1}{c|}{1.54} & \multicolumn{1}{c|}{1.62} & \multicolumn{1}{c|}{1.33} & \multicolumn{1}{c|}{0.56} & \multicolumn{1}{c|}{1.90} & \multicolumn{1}{c|}{27.06} & \multicolumn{1}{c|}{1.56} & \multicolumn{1}{c|}{1.82}  \\ \hline
\multicolumn{1}{|l|}{$\mu$-law} & \multicolumn{1}{c|}{8.69} & \multicolumn{1}{c|}{11.62} & \multicolumn{1}{c|}{8.30} & \multicolumn{1}{c|}{0.41} & \multicolumn{1}{c|}{0.38} & \multicolumn{1}{c|}{78.79} & \multicolumn{1}{c|}{82.02} & \multicolumn{1}{c|}{4.72} & \multicolumn{1}{c|}{12.92} & 1.88 & \multicolumn{1}{c|}{1.84} & \multicolumn{1}{c|}{1.78} & \multicolumn{1}{c|}{1.98} & \multicolumn{1}{c|}{1.62} & \multicolumn{1}{c|}{2.26} & \multicolumn{1}{c|}{1.64} & \multicolumn{1}{c|}{1.62} & \multicolumn{1}{c|}{1.60}  \\ \hline
\end{tabular}
\vspace{-0.2cm}
\end{table*}

%% file: tables/mixed_effects.tex
\begin{table}[tb]
\vspace{-10pt}
\centering
\caption{Model parameters for countermeasure scores with the tested configurations. $\mu$ is the model intercept, $d$ is the class discrimination, $\beta^*$ refers to the biased training effect in the configurations, where $\beta^{\text{spf}} = \beta^*$ and $\beta^{\text{bona}} =  - \beta^*$ and $\varepsilon_{i}$ is the residual variance.}
\renewcommand{\arraystretch}{1.2}
\label{tab:lm_models}
\vspace{-5pt}
\scriptsize
\begin{tabular}{llcccc}
\hline
System & Intervention & \textbf{$\mu$} & d & \textbf{$\beta^*$} & \textbf{$\varepsilon_{i}$} \\ \hline
\multirow{5}{*}{GMM} & MP3 & -0.029 & 0.287 & 0.513 & 0.781 \\
 & Noise & -0.012 & 0.120 & 0.533 & 0.771 \\
 & Loudness & -0.083 & 0.806 & 0.002 & 0.939 \\
 & Nonspeech & -0.025 & 0.243 & 0.341 & 0.901 \\
 & $\mu$-law & -0.056 & 0.549 & 0.173 & 0.948 \\ \hline
\multirow{5}{*}{AASIST-L} & MP3 & -0.064 & 0.623 & 0.379 & 0.849 \\
 & Noise & -0.051 & 0.501 & 0.598 & 0.690 \\
 & Loudness & -0.064 & 0.625 & 0.008 & 0.963 \\
 & Nonspeech & -0.221 & 2.141 & 0.089 & 0.569 \\
 & $\mu$-law & -0.181 & 1.760 & 0.236 & 0.668 \\
\hline
\end{tabular}
\vspace{-0.5cm}
\end{table}

%% file: tables/intervention_snr_blanking.tex
\begin{table}[]
\setlength{\tabcolsep}{20pt}
\renewcommand{\arraystretch}{1.2}
\centering
  \scriptsize
  \caption{Observation model parameters for GMM and AASIST with RawBoost CM for estimated noise and non speech proportion.}
  \vspace{-0.15cm}
  \label{tab:intervention_snr_blanking}
  \setlength{\tabcolsep}{0.3em}
\begin{tabular}{|l|r|r|r|r|}
\hline
&\multicolumn{2}{|c|}{GMM}&\multicolumn{2}{|c|}{AASIST+RawBoost}\\
\hline
\multicolumn{1}{|l|}{Params.} & \multicolumn{1}{|c|}{SNR} & \multicolumn{1}{|c|}{Non speech Prop.} & \multicolumn{1}{|c|}{SNR} & \multicolumn{1}{|c|}{Non speech Prop.}\\ \hline
$\mu$ & -0.95 & -0.30 & -3.87 & -3.39 \\ \hline
$d$ & 2.83 & 1.47& 9.49 & 7.63 \\ \hline
$\beta^{\text{spf}}_\ell$ &0.07&-2.67& -0.07 & -1.49 \\ \hline
$\beta^{\text{bona}}_\ell$ &-0.38&3.78& -0.18 & 3.89 \\ \hline
$\sigma^2$ &2.69&2.66& 2.03 & 2.02 \\ \hline
$\sigma_b^2$ &0.10&0.09& 0.53 & 0.51 \\ \hline
$\sigma_c^2$ &8.57&9.05& 2.71 & 2.71 \\ \hline
Cond. $R^2$ &0.78&0.79& 0.85 & 0.85 \\ \hline
\end{tabular}
\vspace{-0.5cm}
\end{table}

%% file: SECTION_VII_Application_to_ASV.tex
\vspace{-0.25cm}
\section{Case study II: speaker verification}\label{sec:asv-case-study}

Our next task is to consider automatic speaker verification. \textcolor{black}{The relevant consideration for this work is the design of fair ASV performance evaluation, which involves assessing models to ensure their results are reliable and consistent across different data conditions.} While comprehensive treatment on planning and collecting speech data is outside of our scope, we consider the task of \emph{trial design} for an existing speech corpus: given a database of $N$ speech utterances, each assumed to contain one speaker only, the task is to choose a representative subset of trials from the possible $O(N^2)$ comparisons. This is not merely a matter of saving compute but ensuring the reported performance estimates reflect \emph{speaker} comparison. We assume that, in addition to utterance-level speaker IDs, additional demographic and/or recording related metadata is available. How should we design fair evaluation?

The basic principle is to avoid trials that contain potential confounding factors unrelated to speaker traits. Consider first the case of matched speaker IDs across test and reference (enrollment) utterances. The basic rule for these \emph{target} trials is to avoid utterance pairs sourced from the same session \& environment (room, microphone, background). Since the environments are identical and there is no substantial time lapse between the  recording times, a \textit{lazy detector} that compares the spectral characteristics of the environment obtains high 'speaker similarity'. Many of the \emph{speaker recognition evaluation} (SRE) campaigns \cite{doddington2000nist,Greenberg2020-two-decades} organized by \emph{National Institute of Standards and Technology} (NIST) constrain the target trial utterances to have purposefully different telephone handset or codec, environments or language. These choices reduce the possibility that the evaluated system may provide high target scores due to shared environment. For the negative (\emph{nontarget} trials), the mindset is opposite---seeking as comparable conditions across the test and enrollment as possible. Consider an utterance produced by speaker $A$ in a noisy airport, and another utterance produced by speaker $B$ in a silent office. Now, the lazy environment detector produces low score, relating to the combined effects of environment and speaker difference. Another design consideration relates to the demographics of the speakers. In the NIST SREs, it is common to exclude nontarget trials with mismatched genders. Similarly, non-target trials with large age difference can impact performance estimates~\cite{doddington12_odyssey}. By ensuring comparable demographics (matched gender, similar age) one reduced the risk of conflating measurement of speaker comparison.

While the above design principles are well-known in the community, sometimes they are either ignored or misunderstood. One argument is that in applications (e.g. production-level voice biometric systems), one \emph{does encounter} nontargets with gender mismatch and targets with identical mobile devices between enrollment and test. Thus, why exclude them from evaluation? This relates to the different aims of system engineering for specific applications (a short-term goal) vs scientific knowledge discovery (a longer-term goal). For an ASV system that encounters a nontarget identity claim, it may not be important whether the claim is rejected due to 'pure' speaker difference---or combination of speaker and gender and/or device difference. Similarly, as long as a target identity claim is accepted, it may be of little practical relevance whether this happens thanks to detecting speaker traits (or a combination of speaker traits and user's device). Robust scientific discovery, however, demands for generalizable knowledge on speaker comparison that translates to cases where such shortcut cues might be absent. In a sense, one wishes a `pessimistic' trial design that encourages development of systems that can both (i) detect the presence of target speaker under heavily mismatched enrollment and test data; and (ii) where the nontarget speakers could be easily confused with the target speaker due to shared gender, age range, language and other such factors. To make these points clear, we provide demonstations of some more favorable (and less favorable) trial designs on the widely-adopted VoxCeleb dataset~\cite{nagrani2020voxceleb}.

\vspace{-0.25cm}
\subsection{Experimental Setup}
\textcolor{black}{We conducted the ASV experiments using the VoxCeleb dataset, which is widely utilized in the domain of speaker recognition~\cite{nagrani2020voxceleb}.} This dataset is created by extracting audio from the multimedia contents available in YouTube. VoxCeleb dataset is released in two phases named as VoxCeleb 1 and VoxCeleb 2 consisting of the voices of 1251 and 6112 celebrities, respectively. They belong to different genders and nationalities. Even if this dataset is widely used for ASV research, this has several inherent biases. For example, the ASV performance across different speaker groups vary substantially as reported in~\cite{toussaint22_interspeech}. \textcolor{black}{The standard trial lists for evaluating the VoxCeleb also contain some target trials with enrollment and verification utterances from the same YouTube recording session and it may lead to bias.} In our ASV experiments, we have trained the speaker embedding extractor based on state-of-the-art ECAPA~\cite{dawalatabad21_interspeech} model with audio data only from VoxCeleb 2. On the other hand, the ASV trial list consisting enrollment and test utterances was created from the VoxCeleb 1 dataset. The full trial list consists of 290743 target and 290737 non-target trials~\footnote{\url{https://mm.kaist.ac.kr/datasets/voxceleb/meta/list_test_all.txt}}. We have used cosine scoring for similarity measure.

\begin{table}[tb]
\centering
\caption{ASV performance (in terms of EER \%) on VoxCeleb 1 dataset. The flags indicate whether the \emph{recordings} (R), \emph{gender} (G) and \emph{country} (C) for enrollment and verification utterances are same (1) or different (0).}
\scriptsize
\renewcommand{\arraystretch}{1.2}
\label{tab:eer_asv}
\vspace{-5pt}
\begin{tabular}{ccc|ccc|c}
\hline
\multicolumn{3}{c}{\textbf{Target}} & \multicolumn{3}{c}{\textbf{Nontarget}} & \multirow{2}{*}{\textbf{EER}} \\
\textbf{R} & \textbf{G} & \textbf{C} & \textbf{R} & \textbf{G} & \textbf{C} & \\ 
\hline
1 & 1&	1&	0&	0&	0&	\cellcolor{gray!10}0.11\\
1 &	1&	1&	0&	0&	1&	\cellcolor{gray!20}0.13\\
1&	1&	1&	0&	1&	0&	\cellcolor{gray!30}0.24\\
1&	1&	1&	0&	1&	1&	\cellcolor{gray!40}0.39\\
0&	1&	1&	0&	0&	0&	\cellcolor{gray!50}0.69\\
0&	1&	1&	0&	0&	1&	\cellcolor{gray!60}0.83\\
0&	1&	1&	0&	1&	0&	\cellcolor{gray!70}1.59\\
0&	1&	1&	0&	1&	1&	\cellcolor{gray!80}2.63\\
\hline
\end{tabular}
\vspace{-0.5cm}
\end{table}

\textcolor{black}{Table~\ref{tab:eer_asv} shows the ASV performances for different configurations of target and nontarget trials. Unlike the existing work~\cite{toussaint22_interspeech}, where the authors categorized target and non-target trials into four difficulty levels (i.e., trivial, easy, medium, and hard) in absolute terms, our categorization approach is based on the fundamental idea that the difficulty level of the evaluation protocol depends on the combination of target and non-target trials. In Table~\ref{tab:eer_asv}, we list the configurations, starting with the easiest protocol (where target trials originate from the most matched condition and non-target trials from the most mismatched condition), and ending with the most difficult protocol (where target trials originate from the most mismatched condition and non-target trials from the most matched condition). The results demonstrate the importance of trial selection and demonstrates that the ASV performance gradually degrades when the target and nontarget trials are drawn from more similar conditions. Note that the target trials, by definition, belong to the same gender and country. The results reveal that the EER increases from $0.11\%$ to $0.69\%$ when target trials are selected from a different recording, while keeping other factors fixed.}

\textcolor{black}{To further investigate the impact of various data-related factors such as recording, gender, and country, we conduct experiments using an observational approach. For the ASV use case, we focus solely on this approach, as intervening in factors like recording, gender, and country is not feasible.}

\vspace{-0.15cm}
\subsection{Model Description \textcolor{black}{for the Observational Approach in ASV Use Case}}
\vspace{-0.15cm}
We model the association between the trial factors and the score of the classifier. The generic model as defined in~\eqref{eq:gen-LME} can be expressed for the ASV scores as
\vspace{-0.1cm}
\begin{equation}\label{eq:generic-lme-for-vox-data}
     s_{i} = \mu + d,y_i^\text{cls} +  \beta^\text{recording} x_i +\beta^\text{gender} x_i + \beta^\text{country} x_i + {b}_i+\varepsilon_i,
\end{equation}

\vspace{1ex}
    \begin{tabular}{ll}
    where  & \\
        $s_i$               & ASV (main classifier) score for trial $i$ \\
        $y_i^\text{cls}$    & class label ($y_i^\text{cls}=0$, nontarget; $y_i^\text{cls}=1$, target)\\
        $\mu$               & mean score of the nontarget class\\
        $d$                 & class discrimination parameter\\
        $\beta^\text{recording}$ & slope of recording for the bonafide class\\
        $\beta^\text{gender}$ & slope of gender for the spoof class\\
        $\beta^\text{country}$ & slope of country for the spoof class\\
        ${b}_i$ & random per-speaker effect $\text{b}_i \sim \mathcal{N}(0,\sigma_b^2)$\\
        $\varepsilon_i$     & random residual $\varepsilon_i \sim \mathcal{N}(0,\sigma_\varepsilon^2).$ 
    \end{tabular}
    \vspace{-0.2cm}
\vspace{1ex}

\noindent The model \eqref{eq:generic-lme-for-vox-data} parameters are $\vec{\theta}_\text{linreg}=(\mu,d,\beta^\text{recording},\beta^\text{gender},\beta^\text{country},\sigma_\varepsilon^2, \sigma_b^2)$, estimated by fitting the model where the ASV score $s_i$ serves as the output (dependent) variable and the triplets variables describing the trial's class, audio samples as obtained from the same recording, having speech from the same gender, or speakers from the same country are the input (independent) variables. Table \ref{tab:config_models_vox} presents the models conditioned by the class label and the triplets describing recording, gender and country.  The main interest is to identify the differences between the target and nontarget distributions. For example, comparing the difference between the target trials represented by \texttt{(011)} and each of the options for nontarget trials. In this manner, the $\beta$ parameters represent the effect of the specific factor in predicting the ASV score. Additionally, our model contains random effects, a random variable for the variability per-speaker ($b_i$) and a random residual ($\varepsilon_i$) for other variations not described by the predictors included in the model. The speaker effect describes the variability of a random population of speakers.

\input{tables/models_config_vox}

\vspace{-0.25cm}
\subsection{Results and Discussion}
\vspace{-0.15cm}
Table~\ref{tab:params_vox} shows the estimated model parameters. The results indicate that the recording information plays the most important role by showing highest $\beta$ value. The other two factors, i.e., gender and country, exhibit comparable effect indicating that nontarget trials from either opposite gender or different country have similar impact. Higher residual variability compared to the per-speaker variability indicates that the presence of other factors than just within speaker variation. This may be attributed to other session variability effects such as the environmental variations or speaker's age.

\input{tables/lme_vox}

%% file: tables/models_config_vox.tex
\begin{table*}[tb]
\caption{Models per trial class $y_i^\text{cls}$ (0: nontarget, 1: target) for configurations defined by a triplet corresponding to (recording, gender, country) with 1 refering to  \textit{same}. \textit{Difference} refers to the difference of conditional means between the target and nontarget trials distribution.}
\label{tab:config_models_vox}
\vspace{-0.25cm}
\centering
\begin{tabular}{llll}
\hline
$y_i^\text{cls}$ & Config. & Model & \textit{Difference} \\ \hline
1 & \texttt{011}    & $s_i  = \mu + d + \beta^{\text{gender}} + \beta^{\text{country}} + {b}_i+\varepsilon_{i}$ & - \\ \hline
0 & \texttt{000}    & $s_i  = \mu + {b}_i + \varepsilon_{i}$ & $d + \beta^{\text{gender}} + \beta^{\text{country}}$ \\ 
0 & \texttt{001}    & $s_i  = \mu + \beta^{\text{country}}+ {b}_i + \varepsilon_{i}$ & $d + \beta^{\text{gender}}$ \\
0 & \texttt{010}    & $s_i  = \mu + \beta^{\text{gender}} + {b}_i + \varepsilon_{i}$ & $d + \beta^{\text{country}}$ \\
0 & \texttt{011}    & $s_i  = \mu + \beta^{\text{gender}} + \beta^{\text{country}} + {b}_i +\varepsilon_{i}$ & $d$ \\
\hline \hline
1 & \texttt{111}    & $s_i  = \mu + d + \beta^{\text{recording}} + \beta^{\text{gender}} + \beta^{\text{country}} + {b}_i+\varepsilon_{i}$ & -  \\ \hline
0 & \texttt{000}    & $s_i  = \mu + {b}_i+ \varepsilon_{i}$ & $d + \beta^{\text{recording}} + \beta^{\text{gender}} + \beta^{\text{country}}$ \\ 
0 & \texttt{001}    & $s_i  = \mu + \beta^{\text{country}} + {b}_i + \varepsilon_{i}$ & $d+ \beta^{\text{recording}} + \beta^{\text{gender}}$ \\
0 & \texttt{010}    & $s_i  = \mu + \beta^{\text{gender}} + {b}_i + \varepsilon_{i}$ & $d+ \beta^{\text{recording}} + \beta^{\text{country}}$ \\
0 & \texttt{011}    & $s_i  = \mu + \beta^{\text{gender}} + \beta^{\text{country}} + {b}_i + \varepsilon_{i}$ & $d+ \beta^{\text{recording}}$ \\
\hline
     
\end{tabular}%

\vspace{-0.5cm}
\end{table*}

%% file: tables/lme_vox.tex
\begin{table}[h!]
\centering
\caption{Model parameters for speaker verification case.}
\vspace{-0.15cm}
\label{tab:params_vox}
\scriptsize
\begin{tabular}{c c}
\hline
Parameters &  Values \\ \hline
$\mu$ & $0.008$  \\
$d$ & $0.534$  \\
$\beta^{\text{recording}} $ & $0.171$  \\
$\beta^{\text{gender}} $ & $0.009$  \\
$\beta^{\text{country}} $ & $0.011$  \\ 
$\sigma_{{b}_i}^2$ & $0.033^2$ \\
$\sigma_{\varepsilon_{i}}^2$ & $0.106^2$ \\ \hline
$\text{Conditional } R^2$ & 0.881  \\ \hline
\end{tabular}
\vspace{-0.5cm}
\end{table}

%% file: SECTION_DISCUSSION.tex
\vspace{-0.15cm}
\section{Discussion and limitations of the work} \label{sec:discussion}

\subsection{Summary of findings}
\begin{enumerate}
    \item \textcolor{black}{Our interventional study has demonstrated that dataset bias serves as an additional cue for classifier decision-making in anti-spoofing. This advocates for the need of critical evaluation and mitigation strategies to ensure removing biases so that the data, algorithms, and results are reliable in data-driven decision-making processes.}

    \item Our detailed analysis of various types of interventions reveals that introducing perturbations solely to spoof data has a more substantial impact \textcolor{black}{than other configurations.} Out of all the considered cases MP3 coding and noise have higher impact and loudness normalization has lesser impact. \textcolor{black}{These disparities emphasize the complexity of dataset biases and their differing impacts based on data type and applied interventions. A speech dataset is recorded using devices with specific attributes and undergoes further processing through various methods. Spoofed data possibly exhibits greater variation due to differences in spoofing techniques. Understanding these nuances is crucial for developing more effective strategies to address and mitigate bias in datasets, ensuring more reliable and fair outcomes.}
    
    \item For CM tasks, our proposed framework, when applied to observational cases, confirms that the estimated noise characteristics between the two classes differ significantly even on the original data. This difference allows for enhanced class discrimination, particularly when the spoof data undergoes intensification modifications.
    
    \item The observational study with the ASV task reveal that recording information appears to play the most critical role, significantly influencing the model's performance. In contrast, gender and country information have a comparatively lesser impact on the outcomes. This emphasizes more careful target trial design specific to application scenario.
\end{enumerate}

\subsection{Limitations}

While our approach represents a significant step towards systematic analysis of biases in speech-related detection tasks through both interventional and observational approaches, it has been necessary to limit the scope. Our work has a number of limitations that may serve as a basis for future research:
\begin{enumerate}
    \item We focused on detection task, an arguably more fundamental task compared to multi-class tasks, which involves testing multiple individual class-specific hypotheses. While multi-class systems are relevant in applications, this brings up more combinations on the possible  interventions. \textcolor{black}{A rigorous analysis of interventions in a multi-class setting warrants a separate study. The multi-class problem within the proposed framework can be approached in two potential ways. One approach is to decompose the problem into several binary classification tasks, while another is to treat the class information itself as a predictor to explore multidimensional decision scores. This approach sets the stage for future extensions of the framework to a wide range of applications.}
    \item We focused on covariate data shifts. As noted in Section~\ref{subsec:prior-vs-data-shift}, class prevalence shifts are another type of bias requiring separate investigation.
    \item We focused on individual interventions, whereas real-world scenarios involve combination of many factors (e.g. microphone characteristics, reverberation, noise, codec, etc.) Future work may consider combined effects of multiple interventions applied simultaneously.
    \item \textcolor{black}{Our study utilizes two widely used datasets: ASVspoof 2019 LA and VoxCeleb. The overall findings regarding bias align with those reported in existing literature. However, as most studies rely on these same datasets, it would be valuable to conduct experiments with other datasets to determine whether the findings remain consistent or if they too reflect biases stemming from dataset selection.}
    \item Even if our model-free interventional approach is readily applicable to bias analysis in arbitrary detection tasks, in practice, speech expertise was used in defining the interventions. \textcolor{black}{Future studies in other domains will need related expertise to define the relevant interventions and related control parameters. For example, in object recognition, modifications to color, intensity, and background/obstruction serve as interventions, with their degree linked to control parameters. Similarly, in natural language processing, interventions could involve synonyms, irrelevant words, or typographical errors, depending on the nature of the task.}
    \item Finally, we focused solely on \emph{analysis} of shortcut cues. Addressing model training under biased training (or test) data remains a future direction. \textcolor{black}{Additionally, our model assumes that the associated metadata are available for fair evaluation. However, in practice, this assumption may not always hold true, and in such cases, we may need to adopt strategies used in other domains~\cite{hirota2024resampled,jung2022learning,ji2020can}.}
\end{enumerate}

%% file: SECTION_IX_Horse_Arithmetics_Broad_Outlook.tex
\vspace{-0.2cm}
\section{Broader Outlook: Validity of Results in Data-Driven Research}\label{Section:Discussions}
\vspace{-0.1cm}

Before concluding, we like to momentarily step out from the immediate scope of this study to the reader's attention to a broader view on the validity of results in data-driven machine learning. The commentary in this section stems both from interaction with researchers as well as observing general trends in empirical machine learning research. 
We find it instructive to begin from the case of `Clever Hans' \cite{pfungst1911clever}, quoted often as the beginning of research in shortcuts and biased experiments.

\subsection{Can Horses do Arithmetics?}

Can horses count and do arithmetics? Can \emph{a} horse do arithmetics? While we trust that the reader may instinctively answer either \emph{no} or \emph{unlikely}, these seemingly absurd questions puzzled the scientific community in the early 1900s. While the interested reader is encouraged to consult the original source for the amusing story \cite{pfungst1911clever}, \emph{Hans} was a Russian trotting horse trained by Mr. Wilhelm von Osten (1838---1909), a horseman and a former mathematics teacher in a German gymnasium. Von Osten organized free shows in a Berlin neighborhood on a daily basis that appeared to demonstrate abstract conceptual thinking by \emph{Hans}. Among other tasks, \emph{Hans} appeared to be able to count (people, hats, umbrellas, \emph{etc.}), do arithmetics (addition, subtraction, \emph{etc.}), recognize playing cards, name colors, objects, and weekdays of a given year, and even providing preference on pleasant vs. dissonant sounds. 

Due to the apparently entertaining shows reported widely by the German media at the time, the case of \emph{Hans} also raised the interest of the scientific community---albeit with initially widely divided opinions on the underlying mechanism to explain the observed behavior of \emph{Hans}. Perhaps being convinced by the evidence provided by von Osten, and inspired by specific Darwinian theories omnipotent at the time, some researchers were convinced that horses were indeed capable of higher-level conceptual thinking (such as doing arithmetics). Others believed that \emph{Hans} was incapable of conceptual thinking but, rather, had learned the tasks presented to him by heart (note the resemblance to \emph{training data memorization} vs. \emph{generalization} in machine learning). Finally, the third school of researchers believed in \emph{stupid Hans} who neither had memorized the task, nor possessed any high-level cognitive skills, but was instead depending on cues (whether intentional or unintentional) signaled to \emph{Hans} by von Osten. While some were even convinced that there must have been a radiation of `thought-waves' from the brain of von Osten to that of \emph{Hans}, the more plausible hypothesis seemed to be in terms of visual (e.g. hand movements) or auditory cues. These suspections were indeed demonstrated in extensive studies by a specifically-appointed team of scientists, including Oskar Pfungst (1874 -- 1932), a biologist and psychologist who discovered the explanation to be involuntary body language that the trainer communicated to \emph{Hans}. 



Even if the case of \emph{Hans} could be easy to dismiss as being irrelevant to modern machine learning research, it taught an important lesson to the initially-divided scientific community. 
Ever since its discovery, the `\emph{Clever Hans}' effect has had widespread implications for the way how experiments in psychology, and other fields of science are planned and conducted. \textcolor{black}{Besides the applications that we covered in this study, the Clever Hans effect has been demonstrated in various modern machine learning tasks (reviewed in Section \ref{BasicDefinitions}), from music information retrieval \cite{Sturm2014-Horse} and image classification \cite{geirhos2020shortcut} to Alzheimer's disease detection from speech ~\cite{liu24f_interspeech} to name a few. The issues relate to shortcomings in the datasets or experimental protocols. We suspect that there are \emph{many} other datasets that contain (unreported) shortcut cues.} 

\textcolor{black}{From a classification perspective, \emph{the} reason for a shortcuts is a systematic nuisance mechanism correlated with the class label. Shortcuts can inadvertly be introduced, for instance, by compiling the positive and negative class instances from two different data sources (or collected using different pipelines). While it can be practical to compile a benchmark dataset by sampling, for instance, healthy and diagnosed individuals from two different studies, at minimum one should be aware on potential issues in interpreting experimental findings from such datasets. The two data collections or pipelines could differ in various ways, ranging from the selection of the individuals (including their demographics), sensors, environment, filters and the ADC converter, software used, etc, and a classifier may learn to differentiate these background variables. Ideally, the irrelevant variables would be as even as possible across the classes are interested to predict. If it is feasible to control for all such variables at the data collection phase, at minimum \emph{identifying} and \emph{suppressing} (to the extent possible) the most prominent short-cuts through data post-processing should be carried out. Due to justified critiques \cite{muller21_asvspoof,chettri2023clever} on past data collection efforts that the authors have been involved with, we have paid much greater attention to suppressing shortcuts in more recent anti-spoofing data collections~\cite{wang24_asvspoof}. Through these hard lessons, the authors have learnt that collecting data and composing a meaningful evaluation protocol is not as straightforward as it may first seem. We hope this study to raise awareness of shortcuts and the importance of identifying and suppressing them.} 

\textcolor{black}{The interventional methodology that we have put forward in this paper can be seen as a general approach to \emph{simulate} the Clever-Hans effect in a given dataset in a controlled way. Interventional methods themselves are not new (even to speech researchers \cite{chettri2018analysing}), though the intervention types and analysis of results differ drastically between studies; this methodology is not anyhow standardized. In principle, our methodology is general in its applicability to arbitrary black-box binary classifiers. It could be helpful in contrastive analysis of alternative models for their potential reliance on specific shortcuts. The observational methodology, in turn, is applicable to \emph{post-hoc} analysis of short-cuts in a dataset where interventions might be impractical. In both cases, domain knowledge of the researcher is required; in the internventional case, the researcher defines the types of nuisance interventions. In the observational case, the researcher implements the nuisance measurement (e.g. SNR or silence proportion estimation). We expect `fully automated' identification and suppression of shortcuts to require additional, stronger assumptions.}



\subsection{Can DNNs Recognize Objects?}

Can DNNs recognize objects? Can \emph{my} DNN recognize the objects that I trained it to recognize? While we trust that the reader may now instinctively answer \emph{of course}---there are a plethora of convincing real-world applications and we also observe low error rates when we evaluate our models in the laboratory---our intention is to challenge the reader to pause for a moment to think critically the way how we conduct experiments and report results. \textcolor{black}{Besides the issues related to presence of shortcuts, a recurring point emphasized throughout this work (as well as our prior 'warning' study \cite{Hyejin2023-coin-flip} has been issues in \emph{performance evaluation}.}

Whether on the training or test side, \emph{data} forms the basis for all machine learning. To meet the increasing real-world demands, the overall tendency for the past two decades or so has been towards improving performance through extensive experimentation. To this end, we train and evaluate models using datasets of increasing size and complexity ~\cite{zue1990speech, doddington2000nist, nagrani2020voxceleb, wang2020asvspoof} to keep pushing our error rates down. Due to increased demand for reproducibility, 
many researchers publish their data through public data repositories and challenges, allowing alternative modeling solutions to be easily compared. Compared to machine learning research in the past decades where the datasets were either proprietary, small, or poor proxies for tasks outside of the laboratory, the situation has undoubtedly been substantially improved. The general \emph{data-driven} sentiment in deep learning has accelerated practical deployment and opened up completely new applications that would have been impossible in the past.

Despite all these positive prospects, data might have become even self-evident to us that we tend to forget that it might include unwanted built-in biases that lead us astray from the questions we wish to answer. 
In basic science fields, like medicine and social sciences, controlling for potential confounders forms the default way of thinking and is addressed using techniques such as randomized control trials (RCTs); this helps the scientist to isolate superfluous findings from the actual effect of interest (e.g., does a new medicine lead to increased response in specific physiological measures). In contrast, in many empirical machine learning experiments, the researcher often takes the data as-is and trusts it to form a scientifically valid basis for their research, centered around modeling. 
In part due to the large number of parameters in modern DNNs requiring large datasets, the machine learning researcher will typically instead fetch their data (or nowadays often a pre-trained foundational model) from a public repository, whichever is the commonly adopted one in one's domain. The machine learning researcher 
may have no control on the data collection, labeling and validation. Even so, we may tend to think the data (and conclusions drawn upon thereof) as being free of biases. 

\subsection{Does Error Rate of 0\% Indicate a Solved Task?}

Does an error rate of 0\% indicate a solved classification task? By now, we hope the reader is willing to accept at least a possibility that the answer will depend on the task, the classifier, and the data. In the same way as 
\emph{Hans} inspired both the general public and the scientific community 120 years ago, the intelligent-appearing behavior of deep neural networks keeps us undoubtedly inspired these days. Nonetheless, as noted, that convincing behavior of \emph{Hans} also lead the scientific community astray --- away from the important question of the \emph{actual mechanism}. 

Today's machine learning researchers may attend less frequently to shows that display horses doing magic tricks. What we attend to, instead, are numbers associated with specific models, experiments, datasets, evaluation protocols, and evaluation metrics. Sometimes 
the evaluation metric becomes \emph{the} main driving force for trialing out new features, model architectures, loss functions, gradient optimizers, or data augmentation recipes. We 
use 
empirical performance numbers to make statements and conclusions on ideas that worked; what did \emph{not} work; which modeling choices appeared more/less effective. 

To refine the title of this subsection, let us address the following three simple questions (along with answers):
    \begin{enumerate}
        \item \textbf{Is our main response variable (e.g. the score or error rate) an objective, repeatable measurement?} \emph{Yes --- for the specific model and dataset under question}; usually error rates are computed using simple algebraic expressions (such as $N_\text{errors}/N_\text{trials})$ and are repeatable in that sense. But the numbers will change when we change the model and/or the dataset.  
        \item \textbf{Is the performance number reflective of the classification task that we are interested in?} \emph{It depends on the presence of nuisance features and how they may cause an imbalance in the covariates}, as we 
        have argued in this article;
        
        \item \textbf{Is the performance number generalizable to other datasets (e.g. from laboratory to reality)}? \emph{Possibly} --- if the new data happens to miraculously meet the `identically distributed` part in the `i.i.d' assumption often made concerning training-test data relation. In any other situation, the performance reported on a specific dataset should be considered a random outcome 
        dependent on both the classifier and the data distribution.
    \end{enumerate}
Therefore, the answer to the question posed in the title of this subsection title is a clear-cut \textbf{no}. While promising experimental results obtained on challenging datasets and real-life applications can keep all of us inspired, we should refrain from making strong generalizations 
about classification tasks based solely on empirical performance. Given the sheer quantity of experiments conducted these days, machine learning researchers bear a great responsibility to ensure that the reported numbers and findings will be truthful and contribute to science renewal and improved understanding of the modeling tasks. The authors hope that our present study contributes positively at least to awareness of the potential biases in data. 


%% file: SECTION_X_Conclusions.tex
\section{Conclusions}\label{Section:Coclusions}

We propose an approach for linear post-hoc analysis, which aims to unveil the relationship between data-related bias factors and classifier scores. We have developed a straightforward method that introduces diverse interventions during the training and/or testing phases. This approach allows us to address large class of data biases expanding the scope of previous research that was limited to explicit knowledge about the dataset. Our analyses with interventional and observational methods reveals the presence of spurious correlations between data-related factors and classifier scores especially for the domain intensified and domain inverted bias scenario.

%% file: APPENDIX_Nuisance_feature_and_EER.tex
\section*{Appendix: interpretation of the nuisance score}\label{appendix:bias-score-interpretation}


\subsection{Model}

To elaborate on the interpretation of the nuisance score \eqref{eq:perturbation-LLR-computation} and EER values derived from it, consider the simple linear model,
\vspace{-0.2cm}
\begin{equation}\label{eq:structural-eq-for-nuisance-LLR}
        \begin{aligned}
            \ell_i  & = \mu_\ell + d_\ell y_i^\text{cls} + \varepsilon_{i,\ell},   
        \end{aligned}
\end{equation}
where\\
\vspace{2ex} 
    \begin{tabular}{ll}
        $\ell_i$            & Nuisance score for a test instance $x_i$, as in \eqref{eq:perturbation-LLR-computation}\\
        $\mu_\ell$          & Mean nuisance score for the negative class\\
        $d_\ell$            & Nuisance discrimination parameter\\
        $y_i^\text{cls}$    & Class label (either 0 or 1)\\
        $\varepsilon_{i,\ell}$ & Normal residual, $\varepsilon_{i,\ell} \sim \mathcal{N}(0,\sigma_{\varepsilon,\ell}^2).$ 
    \end{tabular}

Equivalently, the class-conditional score distribution are $\ell_i \sim \mathcal{N}(\mu_\ell , \sigma_{\varepsilon,\ell}^2)$ and $\ell_i \sim \mathcal{N}(\mu_\ell + d_\ell, \sigma_{\varepsilon,\ell}^2)$ for the negative and the positive class, respectively. Hence, nuisance scores are modeled as Gaussians with class-specific means but with tied variance. The observed nuisance scores are not necessarily Gaussian, but this simplification facilitates analytic treatment. In specific, $d_\ell$ is interpreted as difference of the class means,
\vspace{-0.2cm}
    \begin{equation}
        d_\ell = \E[\ell_i|y_i^\text{cls}=1]-\E[\ell_i|y_i^\text{cls}=0],         
    \end{equation}
where $\E[\ell_i|\cdot]$ denotes conditional expectation. 
Since $\ell_i$ is constructed from measurements that we prefer a classifier to \emph{not} rely on, from a database designer's perspective, the ideal value is $d_\ell=0$. In this case, the systematic part of the nuisance parameter is not present, so a classifier cannot leverage it.

The unit of $d_\ell$ are LLRs and can take any value in the extended real axis (i.e., including $\pm \infty$). Some readers may find it more intuitive to express the nuisance effect in terms of an error rate instead. For the popular \emph{equal error rate} (EER), $d_\ell$ and EER are functionally related (e.g. \cite{Leeuwen2013-calibrated-LLR,Poh2004-why-do}) through
    \begin{equation}\label{eq:EER-vs-normal-CDF}
        \text{EER}_\mathcal{N}=\Phi(-d_\ell/2),       
    \end{equation}
where $\Phi(\cdot)$ is the cumulative distribution function of the standard normal distribution and where the subscript $\mathcal{N}$ emphasizes this is a model-based EER under the normal distribution assumptions noted. From the properties of $\Phi(\cdot)$, it readily follows
    \begin{itemize}
        \item $d_\ell>0 \Leftrightarrow \text{EER}<0.5$ 
        \item $d_\ell=0 \Leftrightarrow \text{EER}=0.5$ (desired)
        \item $d_\ell<0 \Leftrightarrow \text{EER}>0.5$.
    \end{itemize}
Unlike for the main classifier where the system designer chases for \emph{low} EERs, here the database designer should target at EER of 50\% with the nuisance score. Subject to the assumptions noted in this Appendix, substantial deviations from this reference value are indicative of potential design issues with the dataset. In such cases, the database designer is advised to analyze and mitigate the source of the shortcut, revising the source data and/or training/test partitioning, as needed.


\subsection{Results}


\input{tables/llr_models}
The model \eqref{eq:structural-eq-for-nuisance-LLR} was fitted separately on each of the intervention configurations of Table \ref{tab:config}, for two selected nuisance features (SNR and nonspeech proportion). The parameters   
are shown in Table~\ref{tab:llrmodels}, 
along with empirical EER. 


The results indicate that either nuisance feature could be used to discriminate the classes in both the original, as seen 
from the reasonably high magnitude of $d_l$. 
The impact is more pronounced for SNR. 
For example, $d_\ell=1.430$ in the original data 
indicates that the noise levels in the bona fide and spoof 
classes differ from one another. 



%% file: tables/llr_models.tex
\begin{table}[]
\setlength{\tabcolsep}{20pt}
\renewcommand{\arraystretch}{1.2}
\centering
\scriptsize
\caption{Model parameters for LLR from SNR and non-speech proportion where $\ell$ is SNR and non-speech proportion as a nuisance feature in the defined configurations. The final row shows the empirical EER.}
\vspace{-0.15cm}
\label{tab:llrmodels}
\setlength{\tabcolsep}{0.3em}
\begin{tabular}{|l|c|c|}
\hline
\multicolumn{1}{|l|}{Params.} & \multicolumn{1}{c|}{SNR} & \multicolumn{1}{c|}{Non-speech proportion} \\ \hline
$\mu_\ell$ & -1.094 & 0.278 \\ \hline
$d_\ell$ & 1.430  & 0.256 \\ \hline
${\sigma_\ell}^2$ & 1.739  & 0.162 \\ \hline
Empirical $\mathrm{EER}$ & 34.14 & 20.45 \\ \hline
\end{tabular}
\vspace{-0.2cm}
\end{table}